\documentclass[journal]{IEEEtran}
\usepackage{ifpdf}
\usepackage{amsmath}
\usepackage{amsfonts}
\usepackage{algorithmic}
\usepackage{hyperref}
\usepackage{array}
\usepackage{url}
\usepackage{enumitem}
\usepackage{wrapfig}
\usepackage{cite}
\usepackage{comment}
\usepackage{xcolor}

\newcommand{\update}[1]{\textcolor{black}{#1}}

\DeclareMathOperator*{\argmin}{arg\,min}
\DeclareMathOperator*{\argmax}{arg\,max}

\ifCLASSOPTIONcompsoc
  \usepackage[caption=false,font=normalsize,labelfont=sf,textfont=sf]{subfig}
\else
  \usepackage[caption=false,font=footnotesize]{subfig}
\fi

\ifCLASSINFOpdf
  \usepackage[pdftex]{graphicx}
\else
  \usepackage[dvips]{graphicx}
\fi

\begin{document}
%
\title{Learning a Probabilistic Model for \\ Diffeomorphic Registration}
%
%
%
\author{Julian Krebs, Herv\'{e} Delingette, Boris Mailh\'{e}, Nicholas Ayache and Tommaso Mansi

\thanks{J. Krebs is with the Universit\'{e} C\^{o}te d'Azur, Inria, Epione Team, Sophia Antipolis, 06902 France and also Siemens Healthineers, Digital Services, Digital Technology and Innovation, Princeton, NJ, 08540 USA.}%
\thanks{H. Delingette and N. Ayache are with the Universit\'{e} C\^{o}te d'Azur, Inria, Epione Team, Sophia Antipolis, 06902 France.}%
\thanks{B. Mailh\'{e} and T. Mansi are with Siemens Healthineers, Digital Services, Digital Technology and Innovation, Princeton, NJ, 08540 USA.}%
}



\maketitle

\begin{abstract}
We propose to learn a \update{low-dimensional} probabilistic deformation model from data which can be used for registration and the analysis of deformations. The latent variable model maps similar deformations close to each other \update{in an encoding space. It enables to compare deformations, generate normal or pathological deformations for any new image or to transport deformations from one image pair to any other image.}

Our unsupervised method is based on variational inference. In particular, we use a conditional variational autoencoder (CVAE) network and constrain transformations to be symmetric and diffeomorphic by applying a differentiable exponentiation layer with a symmetric loss function. We also present a formulation that includes spatial regularization such as diffusion-based filters. Additionally, our framework provides multi-scale velocity field estimations. 
We evaluated our method on 3-D intra-subject registration using 334 cardiac cine-MRIs. On this dataset, our method showed state-of-the-art performance with a mean DICE score of 81.2\% and a mean Hausdorff distance of 7.3mm \update{using 32 latent dimensions} compared to \update{three} state-of-the-art methods while also demonstrating more regular deformation fields. The average time per registration was 0.32s. Besides, we visualized the learned latent space and show that the encoded deformations can be used to transport deformations and to cluster diseases with a classification accuracy of 83\% after applying a linear projection.
\end{abstract}


%
\IEEEpeerreviewmaketitle

\section{Introduction}
\IEEEPARstart{D}{eformable} image registration, the process of finding voxel correspondences in a pair of images, is an essential task in medical image analysis \cite{sotiras}. This mapping -- the deformation field -- can be used for example in pre-op / post-op studies, to find the same structures in images from different modalities or to evaluate the progression of a disease. The analysis of geometric changes in successive images is important for instance for diagnosing cardiovascular diseases and selecting the most suited therapies. A possible approach is to register sequential images and analyze the extracted deformations for example by parallel transport \cite{lorenzi2014efficient} or by creating an adapted low-dimensional subspace \cite{rohe2018low}. 

We propose a registration algorithm that learns a deformation model directly from training images. Inspired by recent generative latent variable models, our method learns a low-dimensional probabilistic deformation encoding in an unsupervised fashion. This latent variable space encodes similar deformations close to each other \update{and allows the generation of synthetic deformations for a single image and the comparison and transport of deformations from one case to another}.

\subsection{Deformable Image Registration}
Traditionally, deformable registration is solved by numerical optimization of a similarity metric which measures the distance between the fixed and the deformed moving image. The moving image is warped given a predefined deformation model in order to get closer to the fixed image. Unfortunately, this results in an ill-posed problem which requires further regularization based on prior assumptions \cite{sotiras}. Various regularization energies have been proposed including elastic- \cite{davatzikos1997spatial,burger2013hyperelastic} or diffusion-based methods \cite{thirion1998image,vercauteren2007non,lorenzi2013lcc} (cf.~\cite{sotiras}).  Diffeomorphic transforms are folding-free and invertible. The enforcement of these properties in many medical applications has led to the wide use of diffeomorphic registration algorithms. Popular parametrizations of diffeomorphisms include the Large Deformation Diffeomorphic Metric Mapping (LDDMM) \update{\cite{beg2005computing,cao2005large,zhang2015finite}}, a symmetric normalization approach \cite{avants2008symmetric} or stationary velocity fields (SVF) \cite{arsigny2006log,vercauteren2008symmetric}.

In recent years, learning-based algorithms -- notably Deep Learning (DL) -- have been proposed to avoid long iterative optimization at test time. In general, one can classify these algorithms as supervised and unsupervised. Due to the difficulty of finding ground truth voxel mappings, supervised methods need to rely on predictions from existing algorithms \cite{yang2017quicksilver,rohe2017svf}, simulations \cite{sokooti2017nonrigid,uzunova2017training,eppenhof2018deformable} or a combination of both \cite{krebs2017robust,mahapatra2018deformable}. The latter can be achieved for example by projecting \emph{B-spline} displacement estimations in the space of a statistical deformation model from which one can extract simulations by sampling of its components \cite{krebs2017robust}. Diffeomorphic approaches predict patches of the initial momentum of LDDMMs \cite{yang2017quicksilver} or dense SVFs \cite{rohe2017svf}. Supervised methods are either limited by the performance of existing algorithms or the realism of simulations. Furthermore, retrieving deformations from existing algorithms on a large database is time-consuming and increases the training complexity.

Unsupervised approaches to registration aim to optimize an image similarity, often combined with a penalization or smoothing term (regularization). These learning approaches first appeared in the computer vision community \cite{jason2016back,liang2017dual} and were recently applied to medical image registration \cite{de2017end,balakrishnan2018unsupervised,fan2018birnet,dalca2018unsupervised,tanner2018generative}. Unlike traditional methods, learning-based approaches also \update{can} include task-specific information such as segmentation labels during training while not requiring those labels at test time. Instead of using an image similarity, Hu et al.~\cite{hu2018weakly} proposed to optimize the matching of labels based on a multi-scale DICE loss and a deformation regularization. Fan et al.~\cite{fan2018birnet} proposed to jointly optimize a supervised and unsupervised objective by regressing \emph{ground-truth} deformation fields (from an existing algorithm), while simultaneously optimizing an intensity-based similarity criterion. The disadvantage of these \emph{semi-supervised} approaches is that their training complexity is higher since label information needs to be collected, and for example deformations outside the segmented areas are not guaranteed to be captured. 
Most unsupervised approaches use B-spline grids or dense deformation fields, realized with spatial transformer layers (STN \cite{jaderberg2015spatial}) for an efficient and differentiable linear warping of the moving image. However, it has not been shown yet that these approaches lead to sufficiently regular and plausible deformations. 

\subsection{Deformation Analysis and Transport}
Understanding the deformation or motion of an organ goes beyond the registration of successive images. Therefore, it has been proposed to compare and characterize shape and motion patterns by normalizing deformations in a common reference frame \cite{lorenzi2014efficient,duchateau2011spatiotemporal} and for example by applying statistical methods to study the variation of cardiac shapes \cite{bai2015bi}. \update{In the diffeomorphic setting, various dimensionality reduction methods have been proposed. Vaillant et al.\ \cite{vaillant2004statistics} modeled shape variability by applying PCA in the tangent space to an atlas image. Qiu et al.\ used a shape prior for surface matching \cite{qiu2012principal}. While these methods are based on probabilistic inference, dimensionality reduction is done after the estimation of diffeomorphisms. Instead Zhang et al.\ \cite{zhang2014bayesian} introduced a latent variable model for principle geodesic analysis that estimates a template and principle modes of variation while infering the latent dimensionality from the data. Instead of having a general deformation model capable of explaining the deformations of any image pair in the training data distribution, this registration approach still depends on the estimation of a smooth template. Using the SVF parametrization for cardiac motion analysis,} Roh\'e et al.\ \cite{rohe2018low} proposed to build affine subspaces on a manifold of deformations, the barycentric subspaces, where each point on the manifold represents a 3-D image and the geodesic between two points describes the deformation.

\update{For uncertainty quantification, Wassermann et al.~\cite{wassermann2014probabilistic} used a probabilistic LDDMM approach applying a stochastic differential equation and Wang et al.~\cite{wang2018efficient} employed a low-dimensional Fourier representation of the tangent space of diffeomorphisms with a normal assumption. While both approaches contain probabilistic deformation representations, they have not been used for sampling and the representations have not been learned from a large dataset.}

In the framework of diffeomorphic registration, \emph{parallel transport} is a promising normalization method for the comparison of deformations.  Currently used \emph{parallel transport} approaches are the Schild's \cite{lorenzi2011schild} or pole ladder \cite{lorenzi2014efficient,jia2018parallel} using the SVF parametrization or approaches based on Jacobi fields using the LDDMM parametrization \cite{younes2007jacobi,louis2017parallel}. In general, these approaches aim to convert and apply the temporal deformation of one subject to another subject. However, this \emph{transport} process typically requires multiple registrations, including difficult registrations between different subjects. 

\subsection{\update{Learning-based} Generative Latent Variable Models}
Alternatively and inspired by recently introduced learning-based generative models, we propose to learn a latent variable model that captures deformation characteristics just by providing a large dataset of training images. In the computer vision community, such generative models as Generative adversarial networks (GAN) \cite{goodfellow2014generative}, stochastic variational autoencoders (VAE) \cite{kingma2013auto} and adversarial autoencoders (AAE) \cite{makhzani2015adversarial} have demonstrated great performance in learning data distributions from large image training sets. The learned models can be used to generate new synthetic images, similar to the ones seen during training. In addition, probabilistic VAEs are latent variable models which are able to learn continuous latent variables with intractable posterior probability distributions (encoder). Efficient Bayesian inference can be used to deduce the posterior distribution by enforcing the latent variables to follow a predefined (simple) distribution. Finally, a decoder aims to reconstruct the data from that representation  \cite{kingma2013auto}. As an extension, conditional variational autoencoders (CVAE)  constrain the VAE model on additional information such as labels. This leads to a latent variable space in which similar data points are mapped close to each other. CVAEs are for example used for semi-supervised classification tasks \cite{kingma2014semi}. 

Generative models also showed promising results in medical imaging applications such as in classifying cardiac diseases \cite{biffi2018learning} or predicting PET-derived myelin content maps from multi-modal MRI \cite{wei2018learning}. Recently, unsupervised adversarial training approaches have been proposed for image registration \cite{mahapatra2018deformable,fan2018adversarial,tanner2018generative}. Dalca et al.\ \cite{dalca2018unsupervised} developed a framework which enforces a multivariate Gaussian distribution on each component of the velocity field for measuring uncertainty. However, these approaches do not learn global latent variable models which map similar deformations close to each other in a probabilistic subspace of deformations. To the best of the authors' knowledge, generative approaches for registration which allow the sampling of new deformations based on a learned low-dimensional encoding have not been proposed yet. 

\subsection{Probabilistic Registration using a Generative Model}
We introduce a generative and probabilistic model for diffeomorphic image registration, inspired by generative latent variable models \cite{kingma2013auto,kingma2014semi}. \update{In contrast to other probabilistic approaches such as \cite{yang2017quicksilver,dalca2018unsupervised}}, we learn a low-dimensional global latent space in an encoder-decoder neural network where the deformation of a new image pair is mapped to \update{and where similar deformations are close to each other. This latent space, learned in an unsupervised fashion, can be used to generate an infinite number of new deformations for any single image from the data distribution and not only for a unique template as in the Bayesian inference procedure for model parameter estimation in \cite{zhang2014bayesian}. From this abstract representation of deformations, diffeomorphic deformations are reconstructed by decoding the latent code under the constraint of the moving image. To the best of the author's knowledge, this method describes the first low-dimensional probabilistic latent variable model that can be used for deformation transport from one subject to another. Through applying a latent deformation code of one image pair on a new constraining image, deformation transport (and sampling from the latent space) is useful for instance for simulating cardiac pathologies or synthesizing a large number of pathological and healthy heart deformations for data augmentation purposes.} 

\update{We} use a variational inference method (a CVAE \cite{kingma2014semi}) with the objective of \emph{reconstructing} the fixed image by warping the moving image. The decoder of the CVAE is conditioned on the moving image to ease the \update{encoding task: by making appearance information easily accessible in the decoder (in the form of the moving image), the latent space is more likely to encode deformation rather than appearance information. This implicit decoupling of deformation and appearance information allows to transport deformations from one case to another by pairing a latent code with a new conditioning image.} The framework provides multi-scale estimations where velocities are extracted at each scale of the decoding network. We use the SVF parametrization and diffeomorphisms are extracted using a vector field exponentiation layer, based on the \emph{scaling and squaring} algorithm proposed in \cite{arsigny2006log}. This algorithm has been successfully applied in neural networks in our previous work \cite{krebs2018unsupervised} and in \cite{dalca2018unsupervised}. The framework contains a dense spatial transformer layer (STN) and can be trained \update{end-to-end} with a choice of similarity metrics: to avoid asymmetry, we use a symmetric and normalized local cross correlation criterion. In addition, we provide a generic formulation to include regularization terms to control the deformation appearance (if required), such as diffusion regularization in form of Gaussian smoothing \cite{lorenzi2013lcc}. During training, similarity loss terms for each scale and a loss term enforcing a prior assumption on the latent variable distribution are optimized by using the concept of \emph{deep supervision} (cf.\ \cite{lee2015deeply}). During testing, \update{the low-dimensional} latent encoding, multi-scale \update{estimations of velocities, deformation fields} and deformed moving image are retrieved in a single forward path of the neural network. 

We evaluate our framework on the registration of cardiac MRIs between end-diastole (ED) and end-systole (ES) and provide an intensive analysis on the structure of the latent code and evaluate its application for transporting encoded deformations from one case to another. 

This paper extends our preliminary work \cite{krebs2018unsupervised} by adding:
\begin{itemize}[noitemsep]
\item Detailed derivations of the probabilistic registration framework including a generic regularization model.
\item Deep supervision, multi-scale estimations and a normalized loss function to improve registration performances.
\item Analysis of size and structure of the latent variable space.
\item Evaluation of the deformation transport by comparing it to a state-of-the-art algorithm \cite{lorenzi2014efficient}.
\end{itemize}

\section{Methods}
In image registration, the goal is to find the spatial transformation $\mathcal{T}_z:\mathbb{R}^3\rightarrow\mathbb{R}^3$ which is parametrized by a $d$-dimensional vector $z\in\mathbb{R}^d$. The optimal values of $z$ are the ones which best warp the moving image $M$ in order to match the fixed image $F$ given the transformation $\mathcal{T}_z$. Both images $F$ and $M$ are defined in the spatial domain $\Omega\in\mathbb{R}^3$. Typically, $z$ is optimized by minimizing an objective function of the form: $\argmin_z  \mathcal{F}(z,M,F)= \mathcal{D}\left(F,M \circ\ \mathcal{T}_z) + \mathcal{R}(\mathcal{T}_z\right)$, where $\mathcal{D}$ is a metric measuring the similarity between fixed $F$ and warped moving image $M \circ\ \mathcal{T}_z$. $\mathcal{R}$ is a spatial regularizer \cite{sotiras}. Recent unsupervised DL-based approaches \cite{jason2016back,de2017end,balakrishnan2018unsupervised} try to learn to maximize such a similarity metric $\mathcal{D}$ using stochastic gradient descent methods and a spatial transformer layer (STN \cite{jaderberg2015spatial}) for warping the moving image $M$. 

In extension, we propose to model registration by learning a probabilistic deformation parametrization vector $z$ from a set of example image pairs $(M, F)$. Thereby, we constrain the low-dimensional $z$ to follow a prior distribution $p(z)$. In other words, our approach contains two key parts: a latent space encoding to model deformations and a decoding function that aims to \emph{reconstruct} the fixed image $F$ from this encoded transformation -- by warping the moving image $M$. In addition, this decoding function is generative as it allows to sample new deformations based on $p(z)$.

\subsection{Probabilistic model for multi-scale registration}
\begin{figure}[tb]
\centering 
\subfloat[]{\includegraphics[trim=490 280 167 46,clip,width=0.5\linewidth]{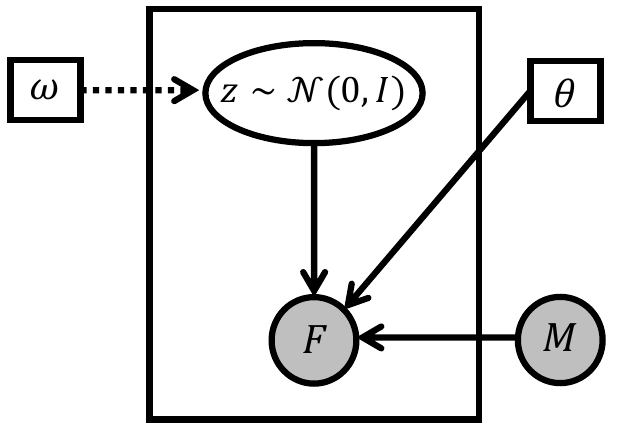}\label{stoch_proc_figure}}\hfill
\subfloat[]{\includegraphics[trim=35 225 620 46,clip,width=0.5\linewidth]{figures/stoch_proc.pdf}\label{stoch_proc_reg_figure}}
\caption{\small{(a) Generative process for registration representing the likelihood of the fixed image $F$ given the latent variable vector $z$ and moving $M$: $p_\theta(F|z,M)$, where $\omega$ and $\theta$ are fixed parameters. (b) Generative process for regularized image registration where the likelihood depends on the regularized velocities $p_\theta(F|v^*,M)$.}}
\end{figure}

We assume a generative probabilistic distribution for registration $p_{true}(F|M)$, capturing the deformation from $M$ towards $F$. We aim at learning a parameterized model $p_\theta(F|M)$ with parameters $\theta$ which allows us to sample new $F$'s that are similar to samples from the unknown distribution $p_{true}$. To estimate the posterior $p_\theta$ we use a latent variable model parameterized by $z$. 
Following the methodology of a VAE \cite{kingma2013auto}, we assume the prior $p(z)$ to be a multivariate unit Gaussian distribution with spherical covariance $I$:
\begin{equation}\label{p_z}
p(z) \sim \mathcal{N}(0, I).
\end{equation}
Using multivariate Gaussians offers a closed-form differentiable solution, however, $p(z)$ could take the form of other distributions. In this work, we parameterize deformation fields $\phi$ by stationary velocity fields (SVF), denoted by velocities $v$: $\phi=\text{exp}(v)$ \cite{arsigny2006log}. \update{These transformation maps $\phi$ are given as the sum of identity and displacements $u(x)$ for every position $x\in\Omega$: $\phi(x)=x+u(x)$}. In the multi-scale approach, we define velocities $v^s$ at scale $s\in \mathcal{S}$ where $\mathcal{S}$ is the set of different image scales ($s=1$ describes the original scale for which we omit writing $s$ and $s=2,3,...$ the scale, down-sampled by a factor of $2^{s-1}$). For each scale $s$, a family of \emph{decoding} functions $f_v^s$ is defined, parameterized by a fixed $\theta^s \subset \theta$ and dependent on $z$ and the moving image $M^s$: 
\begin{equation}\label{f_v}
v^s = f_v^s(z, M^s; \theta^s).
\end{equation}
In the training, the goal is to optimize $\theta^s$ such that all velocities $v^s$ are likely to lead to warped moving images $M^{*s}$ that will be similar to $F^s$ in the training database.  $M^{*s}$ is obtained by exponentiation of $v^s$ and warping of the moving image. Using Eq.\ \ref{f_v}, we can define the families of functions $f^s$: 
\begin{equation}\label{f_}
M^{*s} := f^s(z, M^s; \theta^s) = M^s\circ \text{exp}(f_v^s(z, M^s; \theta^s)).
\end{equation}
In order to express the dependency of $f^s$ on $z$ and $M^s$  explicitly, we can define a distribution $p(F^s | z,M^s;\theta^s)$. The product over the different scales gives us the output distribution: 
\begin{equation}\label{outputdist}
p_\theta(F | z,M) = \prod_{s\in \mathcal{S}} p(F^s | z,M^s;\theta^s).
\end{equation}
By using the law of total probability, this leads to the following stochastic process for computing $p_\theta(F|M)$ which is also visualized in Fig.~\ref{stoch_proc_figure} (cf.\ \cite{kingma2014semi}):
\begin{equation}\label{stochPrc}
p_\theta(F|M) = \int_z p_\theta(F | z,M) p(z) dz.
\end{equation}
The likelihood $p_\theta(F | z,M)$ can be any distribution that is computable and continuous in $\theta$. In VAEs, the choice is often Gaussian, which is equivalent to adopting a sum-of-squared differences (SSD) criterion (cf.\ \cite{kingma2013auto}). We propose instead to use a local cross-correlation (LCC) distribution due to its robustness properties and superior results in image registration compared to SSD (cf.\ \cite{lorenzi2013lcc,avants2011reproducible}). Thus, we use the following Boltzmann distribution as likelihood:
\begin{equation}\label{lcc}
p_\theta^{s}(F^s | z,M^s) \sim \text{exp}(-\lambda \mathcal{D}_{LCC}(F^s,M^s,v^s)),
\end{equation}
where $v^s=f^s_{v}(z, M^s; \theta^s)$ are the velocities and $\lambda$ is a scalar hyperparameter. The symmetric $\mathcal{D}_{LCC}$ is defined as: 
\begin{multline}\label{normLCCDef}
\mathcal{D}_\textit{LCC}(F^s,M^s,v^s) = \\ \frac{1}{P} \sum_{x \in \Omega}  \frac{\sum_i \left(\left(F^{*s}_{x_i} - \overline{F^{*s}_x}\right)\left(M^{*s}_{x_i} - \overline{M^{*s}_x}\right)\right)^2}{\left(\sum_i \left(F^{*s}_{x_i} - \overline{F^{*s}_x}\right)^2\right) \left(\sum_i \left(M^{*s}_{x_i} - \overline{M^{*s}_x}\right)^2\right) + \tau} - 1,
\end{multline}
with $P$ pixels $x\in \Omega$, the symmetrically warped images $M^{*s}=M^s\circ \text{exp}\left(v^s/2\right)$ and $F^{*s}=F^s\circ \text{exp}\left(-v^s/2\right)$. The bar $\overline{F_x}$ symbolizes the local mean grey levels of $F_x$ derived by mean filtering with kernel size $k$ at position $x$. $i$ is iterating through this $k\times k$-window. A small constant $\tau$ is added for numerical stability ($\tau=1e^{-15}$). 

\subsubsection{Learning the constrained deformation encoding}
\update{In order to optimize the parameterized model over $\theta$ (Eq.~\ref{stochPrc}),} two problems must be solved: First, how to define the latent variables $z$, for example decide what information these variables represent. VAEs assume there is no simple interpretation of the dimensions of $z$ but instead assert that samples of $z$ are drawn from a simple distribution $p(z)$. 

Second, the integral over $z$ is intractable since one would need to sample a too large number of $z$'s to get an accurate estimate of $p_\theta(F|M)$. Instead of sampling a large number of $z$'s, the key assumption behind VAEs is to sample only $z$'s that are likely to have produced $F$ and compute $p_\theta(F|M)$ only from those. To this end, one needs to compute the intractable posterior $p(z|F,M)$. Due to this intractability, in VAEs \cite{kingma2013auto}, the posterior is approximated by learning an \emph{encoding} distribution $q_\omega(z|F,M)$, using a neural network with parameters $\omega$ (the encoder). This approximated distribution can be related to the true posterior using the Kullback-Leibler divergence (KL) which leads (after rearranging the terms) to the evidence lower bound (ELBO) of the log marginalized likelihood $\text{log } p_\theta(F|M)$ (cf.~\cite{kingma2013auto,kingma2014semi}):
\begin{multline}\label{objective}
\text{log } p_\theta(F|M) - \text{KL}\left[q_\omega(z|F,M)\| p(z|F,M) \right] = \\ 
\mathbb{E}_{z\sim q}\Big[\text{log } p_\theta(F|z,M) \Big] - \text{KL}\left[q_\omega(z|F,M)\| p(z) \right].
\end{multline}
The KL-divergence on the left hand side gets smaller the better $q_\omega(z|F,M)$ approximates $p(z|F,M)$ and ideally vanishes if $q_\omega$ is of enough capacity. Thus, maximizing $\text{log } p_\theta(F|M)$ \update{is equivalent} to maximizing the ELBO on the right hand side of the equation consisting of encoder $q_\omega$ and decoder $p_\theta$ which can be both optimized via stochastic gradient descent.

\subsubsection{Optimizing the \update{ELBO}}
\update{According to the right-hand side of Eq.~\ref{objective}, there are two terms} to optimize, the KL-divergence of prior $p(z)$ and encoder distribution $q_\omega(z|F,M)$ and the expectation of the reconstruction term $\text{log } p_\theta(F|z,M)$.
Since the prior is a multivariate Gaussian, the encoder distribution is defined as $q_\omega(z|F,M)=\mathcal{N}(z|\mu_\omega(F,M),\Sigma_\omega(F,M))$, where $\mu_\omega$ and $\Sigma_\omega$ are deterministic functions learned in an encoder neural network with parameters $\omega$. The KL-term can be computed in closed form as follows (constraining $\Sigma_\omega$ to be diagonal):
\begin{multline*}
\text{KL}[\mathcal{N}(\mu_\omega(F,M), \Sigma_\omega(F,M))\| \mathcal{N}(0,I)] = \\
\frac{1}{2} \Big( \text{tr}(\Sigma_\omega(F,M)) + \|\mu_\omega(F,M) \| - k - \text{log det} (\Sigma_\omega(F,M)) \Big),
\end{multline*}
where $k$ is the dimensionality of the distribution. 

\update{The expected log-likelihood $\mathbb{E}_{z\sim q}\left[\text{log } p_\theta(F|z,M) \right]$, the reconstruction term, could be estimated by using many samples of $z$. To save computations, we treat $p_\theta(F|z,M)$ as $\mathbb{E}_{z\sim q}\left[\text{log } p_\theta(F|z,M) \right]$ by only taking one sample of $z$. This can be justified as optimization is already done via stochastic gradient descent, where we sample many image pairs $(F,M)$ from the dataset $\mathcal{X}$ and thus witness different values for $z$. This can be formalized with the expectation over $F,M\sim \mathcal{X}$:}
\begin{equation*}
\update{
\mathbb{E}_{F,M\sim \cal{X}} \bigg[ \mathbb{E}_{z \sim q} \Big[\text{log } p_{\theta}(F|z,M) \Big] - \text{KL}\left[q_\omega(z|F,M)\| p(z) \right] \bigg].}
\end{equation*}
\update{To enable back-propagation through the sampling operation $q_\omega(z|F,M)$, the \emph{reparametrization} trick \cite{kingma2013auto} is used in practice, where $z=\mu_\omega + \epsilon \Sigma_\omega^{1/2}$ (with $\epsilon \sim \mathbf{N}(0,I)$)}. Thus, for image pairs $(F,M)$ from a training dataset $\cal{X}$ the actual objective becomes:
\update{\begin{multline}
\mathbb{E}_{F,M\sim \cal{X}} \bigg[ \mathbb{E}_{\epsilon \sim \mathcal{N}(0,I)} \Big[\text{log } p_{\theta}(F|z=\mu_\omega(F,M) + \\ \Sigma_\omega^{1/2}(F,M) \ast \epsilon,M) \Big] - 
\text{KL}\left[q_\omega(z|F,M)\| p(z) \right] \bigg].
\end{multline}}
\update{After insertion of Eq.~\ref{outputdist}, the $\text{log}$ of the product over the scales $s\in \mathcal{S}$ results in the sum of the log-likelihood distributions:}
\begin{multline}\label{repara_trick}
\mathbb{E}_{F,M\sim \cal{X}} \bigg[ \mathbb{E}_{\epsilon \sim \mathcal{N}(0,I)} \Big[\sum_{s\in \mathcal{S}} \text{log } p_{\theta^s}(F^s|z=\mu_\omega(F,M) + \\ \Sigma_\omega^{1/2}(F,M) \ast \epsilon,M^s) \Big] - 
\text{KL}\left[q_\omega(z|F,M)\| p(z) \right] \bigg].
\end{multline}

\begin{figure*}[tb]
\centering 
\subfloat[]{\includegraphics[trim=4 156 415 5,clip,width=.60\linewidth]{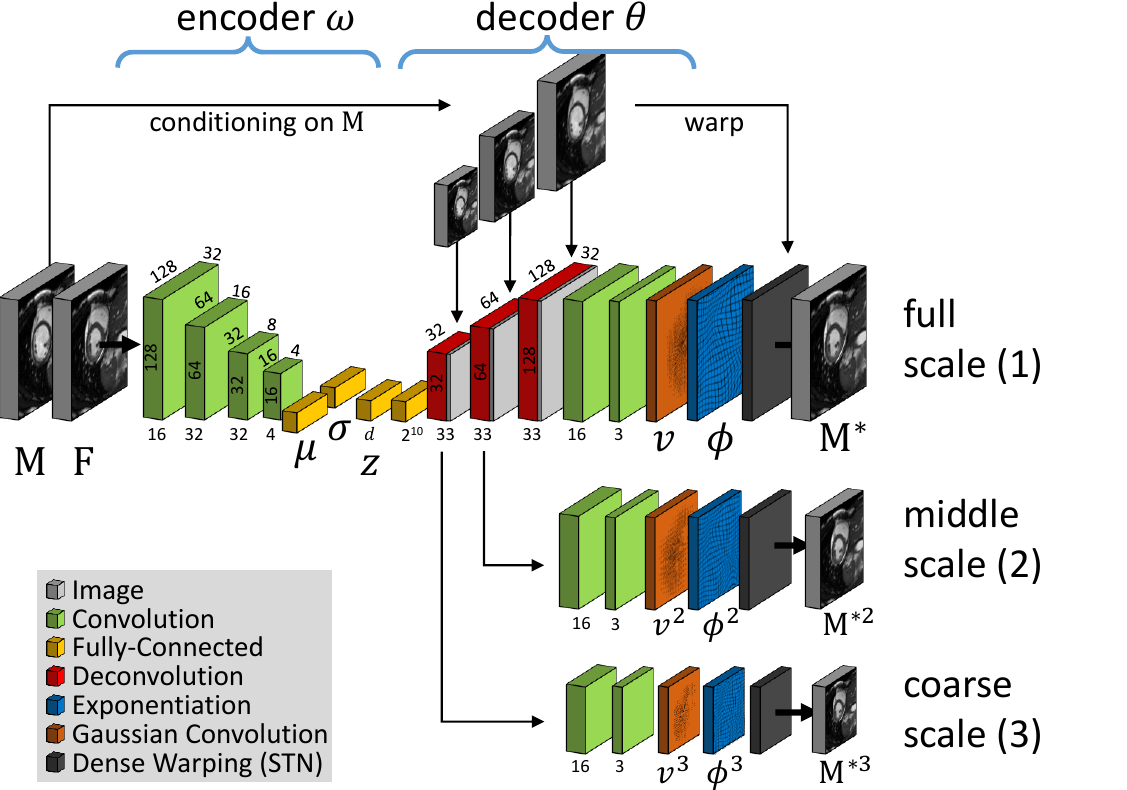}\label{architecture}}\hfill
\subfloat[]{\includegraphics[trim=5 255 657 2,clip,width=.36\linewidth]{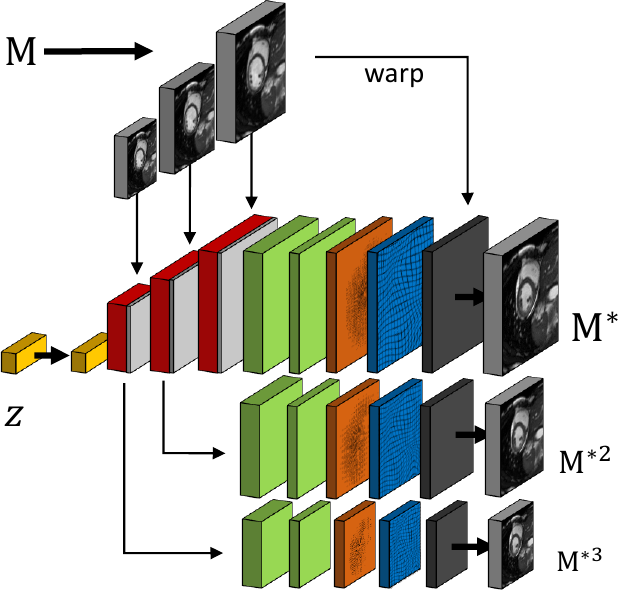}\label{decoder}}
\caption{\small{(a) Probabilistic multi-scale registration network based on a CVAE. An encoder maps deformations \update{to latent variables $z\in\mathbb{R}^d$ (with for example $d=32$)} from which a decoder extracts velocities and diffeomorphisms at different scales while being conditioned on the moving image $M$. (b) After training, the decoder network can be also used to sample and transport deformations: Apply $z$-code on any new image $M$. 
}}
\end{figure*}

\subsection{Introducing regularization on velocities}
So far, we have considered that at each scale $s$, a velocity field $v^s$ is generated by a decoding function $f_v^s(z,M^s;\theta^s)$ through a neural network. To have a better control of its smoothness, we propose to regularize spatially $v^s$ through a Gaussian convolution with standard deviation $\sigma_G$: 
\begin{equation}\label{gaussianfilter}
\hat{v}^s = G_{\sigma_G} \ast v^s
\end{equation}

Gaussian smoothing  was applied here, but it could be replaced by any quadratic Tikhonov regularizer or by any  functional enforcing prior knowledge about the velocity field.

In the remainder, we show how the regularization of velocities can be inserted into the proposed probabilistic framework.  To make the notation less cluttered, we drop the scale $s$ superscript in the velocity notations.  Until now, the velocities $v$ have been handled as fixed parameters $v=f_v(z,M;\theta)$. We can equivalently assume that velocities $v$ are random variables with a Dirac posterior probability : $p_\theta(v|z,M) \sim \delta_{f_v(z,M;\theta)} (v)$. 

We now introduce the random variable $v^*$ which represents the regularized velocities as shown in Fig.~\ref{stoch_proc_reg_figure}. This quantity is linked to  the  regular velocities $v$ through a Gaussian distribution $p(v|v^*)=G(v^*,1)$ such that $v$ is close to $v^*$ in terms of $L_2$ norm. Furthermore, we define a  diffusion-like regularization prior on $v^*$ \cite{nielsen1997regularization}:
\begin{equation*}
\text{log} \hspace{1mm} p(v^*) \propto \int_\Omega \sum_{i=1}^\infty \update{\frac{\sigma_G^{2i}}{2^i i!}} \left[ \frac{\partial^i}{\partial \Omega^i} v^*\right]^2 \hspace{1mm} d\Omega,
\end{equation*}
\update{taking into account the Taylor expansion of the Fourier transform of the Gaussian.} The maximum {\em a posteriori} of the regularized velocities  $\hat{v}$ is then obtained through Bayes law:
\begin{equation}
\hat{v} = \argmax_{v^*} \mathrm{log} ~ p(v^*|v) =\argmax_{v^*} \mathrm{log} p(v|v^*) +  \mathrm{log} p(v^*)\nonumber
\end{equation}
which in this case is equivalent to solving the Heat equation~\cite{nielsen1997regularization} and leads to a Gaussian convolution: $\hat{v} = G_{\sigma_G} \ast v$.

Finally, we conveniently assume that the posterior probability of $v^*$ is infinitely peaked around its mode, {\em i.e.} $p(v^*|\update{v}) \sim \delta_{\hat{v}} (v^*)$ (\update{assumption sometimes made} for the Expectation-Maximization algorithm~\cite{DBLP:journals/neco/KuriharaW09}). \update{In the decoding process, we can now marginalize out the velocity variables $v$ and $v^*$ by integrating over  both such that only $\hat{v}$ remains:}
\begin{align}\label{stochprocreg}
p_\theta(F|M) &= \int_z \int_v \int_{v^*} p(F|v^*,M)~p(v^*|v)~p_\theta(v|z,M)~p(z)\nonumber\\& \hspace{160pt} dv~ dv^*~dz \nonumber\\
&= \int_z  p(F|\hat{v},M)~p(z)~dz.
\end{align}
Thus, the proposed graphical model leads to a decoder working with the regularized velocity field $\hat{v}$ instead of the $v$ generated by the neural network. 
When combining regularized velocities $\hat{v}^s$ at all scales, we get: 
\begin{equation}
p_\theta(F|M) = \int_z \prod_{s \in \mathcal{S}} p(F^s|\hat{v}^{s},M^s)~p(z)~dz.
\end{equation}
This can be optimized as before and leads to Gaussian convolutions at each scale if considering diffusion-like regularization. Thus, the multi-scale loss function per training image pair ($F$,$M$) for one sample $\epsilon$ is defined as (cf.~ Eq.~\ref{repara_trick}):
\begin{multline}\label{final_loss}
 \argmin_{\omega,\theta} \frac{1}{2} \Big( \text{tr}(\Sigma_\omega) + \mu_\omega^\top \mu_\omega - k - \text{log det} (\Sigma_\omega) \Big) \\ - \lambda \sum_{s \in \mathcal{S}} \mathcal{D}_\textit{LCC}(F^s,M^s,\hat{v}^{s}),
\end{multline}
where $\hat{v}^s$ depends on $v^s$ and therefore on $\theta$ (cf.~Eq.~\ref{f_v} and \ref{gaussianfilter}).

\subsection{Network architecture}
The encoder-decoder neural network takes the moving and the fixed image as input and outputs the latent code $z$, velocities $v$, the deformation field $\phi$ and the warped moving image $M^*$. The last three are returned at the different scales $s$. The encoder consists of strided convolutions while the bottleneck layers ($\mu$, $\sigma$, $z$) are fully-connected. The deconvolution layers in the decoder were conditioned by concatenating each layer's output with sub-sampled versions of $\mathbf{M}$. \update{Making appearance information of the moving image easily accessible for the decoder, allows the network to focus on deformation information -- the differences between moving and fixed image -- that need to pass through the latent bottleneck. While it is not guaranteed that the latent representation contains any appearance information, it comes at a cost to use the \emph{small} bottleneck for appearance information.} At each decoding scale, a convolutional layer reduces the number of filter maps to three. Then, a Gaussian smoothing layer (cf.\ Eq.\ \ref{gaussianfilter}) with variance $\sigma_G^2$ is applied on these filter maps. The resulting velocities $v^s$ (a SVF) are exponentiated by the \emph{scaling and squaring} layer \cite{krebs2018unsupervised} in order to retrieve the diffeomorphism $\phi^s$ which is used by a dense STN to retrieve the warped image $M^{*s}$. The latent code $z$ is computed according to the reparametrization trick. During training, the network parameters are updated through back-propagation of the gradients with respect to the objective Eq.\ \ref{repara_trick}, defined at each multi-scale output. Finally during testing, registration is done in a single forward path where $z$ is set to $\mu$ since we want to execute registration deterministically. One can also think of drawing several $z$ using $\sigma$ and use the different outputs for uncertainty estimation as in \cite{dalca2018unsupervised} which we do not further pursue in this work. The network architecture can be seen in Fig.\ \ref{architecture}. Besides registration, the trained probabilistic framework can be also used for the sampling of deformations as shown in Fig. \ref{decoder}. 

\section{Experiments}
We evaluate our framework on cardiac intra-subject registration. End-diastole (ED) frames are registered to end-systole (ES) frames from cine-MRI of healthy and pathological subjects\update{. These images show large deformations.} Additionally, we evaluate the learned encoding of deformations by visualizing the latent space and transporting encoded deformations from one patient to another. All experiments are in 3-D. 

\begin{figure*}[tb]
\centering 
\includegraphics[trim=1 308 53 0,clip,width=1.\linewidth]{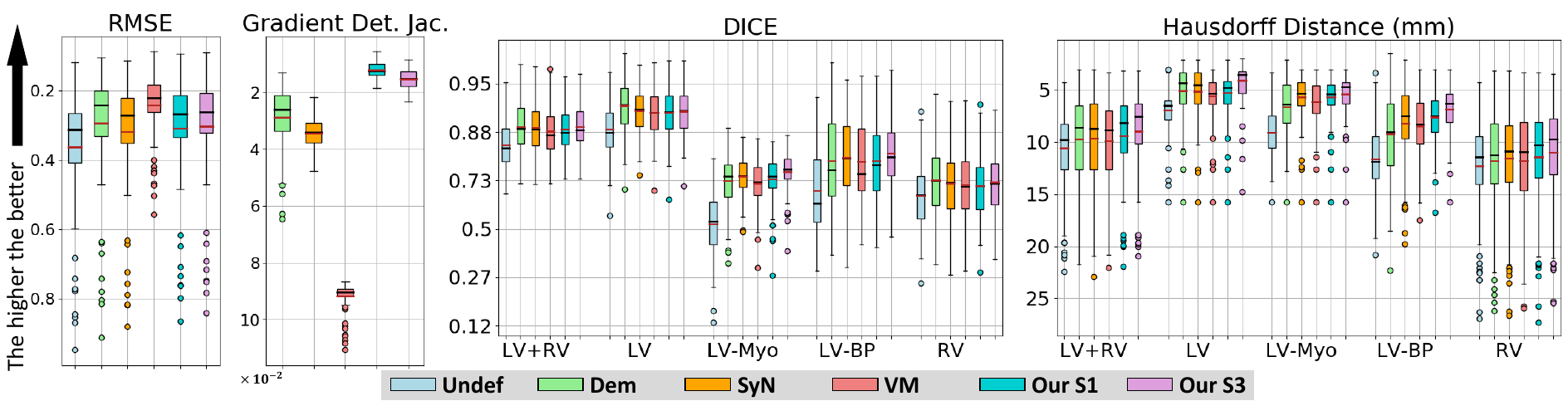}
\caption{\small{Boxplots of registration results comparing the undeformed (Undef) case to the different algorithms: lcc-demons (Dem), \update{SyN}, voxelmorph (VM) and our single scale (S1) respectively multi-scale (S3) using \update{RMSE, gradient of the determinant of the Jacobian, DICE scores (logit-transform) and Hausdorff distances (HD in mm)}. Mean values are denoted by red bars. Higher values are better.}}\label{cardiac_metric}
\end{figure*}

\subsubsection{Data} We used the 334 ED-ES frame pairs of short-axis cine-MRI sequences. 184 cases were acquired from different hospitals and 150 cases were used from the Automatic Cardiac Diagnosis Challenge (ACDC) at STACOM 2017 \cite{bernard2018deep}, mixing congenital heart diseases with images from adults. We used 234 cases for training and for testing 100 cases from ACDC, that contain segmentation and disease class information. The testing set contained 20 cases of each of the following cardiac diseases: dilated cardiomyopathy (DCM), hypertrophic cardiomyopathy (HCM), previous myocardical infarction (MINF), abnormal right ventricle (RV) and healthy (Normal). All images were resampled with a spacing of $1.5 \times 1.5 \times 3.15$ mm and cropped to a size of $128\times128\times32$ voxels, by equally removing voxels from all sides. These dimensions were chosen to save computation time and are not a limitation of the framework.

\subsubsection{Implementation details} Our neural network consisted of four encoding convolutional layers with strides (2, 2, 2, 1) and three decoding deconvolutional layers. Each scale contained two convolutional layers and a convolutional Gaussian layer with $\sigma_G=3mm$ (kernel size 15) in front of an exponentiation and a spatial transformer layer using trilinear interpolation (cf.~Fig.\ \ref{architecture}). The dimensionality of the latent code $z$ was set to $d=32$ as a compromise of registration quality and generalizability (cf.\ experiment on latent vector dimensionality). The number of trainable parameters in the network was $\sim$420k. LeakyReLu activation functions and L2 weight decay of $1*10^{-4}$ were applied on all layers except the last convolutional layer in each scale where a \emph{tanh} activation function was used in order to avoid extreme velocity values during training. All scales were trained together, using linearly down-sampled versions of the input images for the coarser scales. In all experiments, the number of iterations in the exponentiation layer was set to $N=4$  (evaluated on a few training samples according to the formula in \cite{arsigny2006log}). During the training, the mean filter size of the LCC criterion was $k=9$. The loss hyper parameter was empirically chosen as $\lambda=5000$ such that the similarity loss was optimized while the latent codes roughly had zero means and variances of one. We applied a learning rate of $1.5*10^{-4}$ with the Adam optimizer and a batch size of one. For augmentation purposes, training image were randomly shifted, rotated, scaled and mirrored. The framework has been implemented in \textit{Tensorflow} using \textit{Keras}\footnote{https://keras.io/}. Training took $\sim$24 hours and testing a single registration case took 0.32s on a \textit{NVIDIA GTX TITAN X} GPU. 

\subsubsection{Registration}
We compare our approach with the LCC-demons (Dem, \cite{lorenzi2013lcc}) \update{and the ANTs software package using Symmetric Normalization (SyN, \cite{avants2008symmetric})} with manually tuned parameters (on a few training images) and the diffeomorphic DL-based method VoxelMorph \cite{dalca2018unsupervised} (VM) which has been trained using the same augmentation techniques as our algorithm. For the latter, we set $\sigma=0.05$, $\lambda=50000$ and applied a reduced learning rate of $5*10^{-5}$ for stability reasons while using more training epochs. Higher values for $\lambda$ led to worse registration accuracy. We also show the improvement of using a multi-scale approach (with 3 scales, S3) compared to a single-scale objective (S1). 

\begin{wrapfigure}[11]{r}{0.37\columnwidth}
\centering 
\vspace{-6pt}
\includegraphics[trim=0 270 660 0,clip,width=0.36\columnwidth]{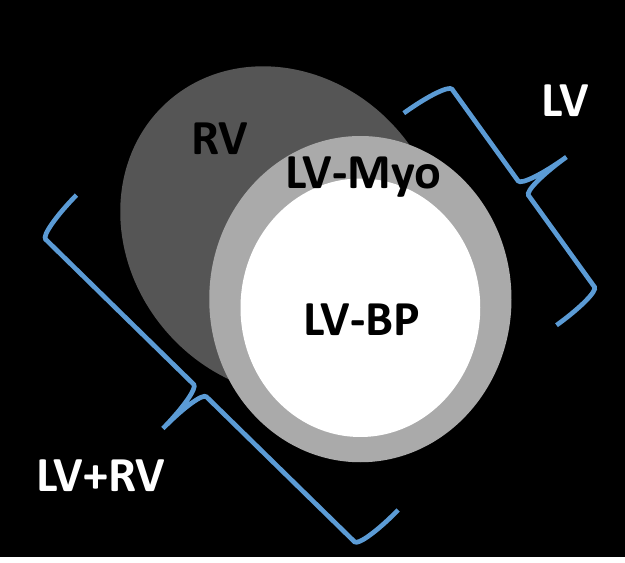}
\vspace{-17pt}
\caption{\small{Cardiac structures used only for measuring registration accuracy.}}\label{masks}
\end{wrapfigure}

\begin{figure*}[tb]
\centering 
\subfloat[]{\includegraphics[trim=2 124 294 0,clip,width=1.\linewidth]{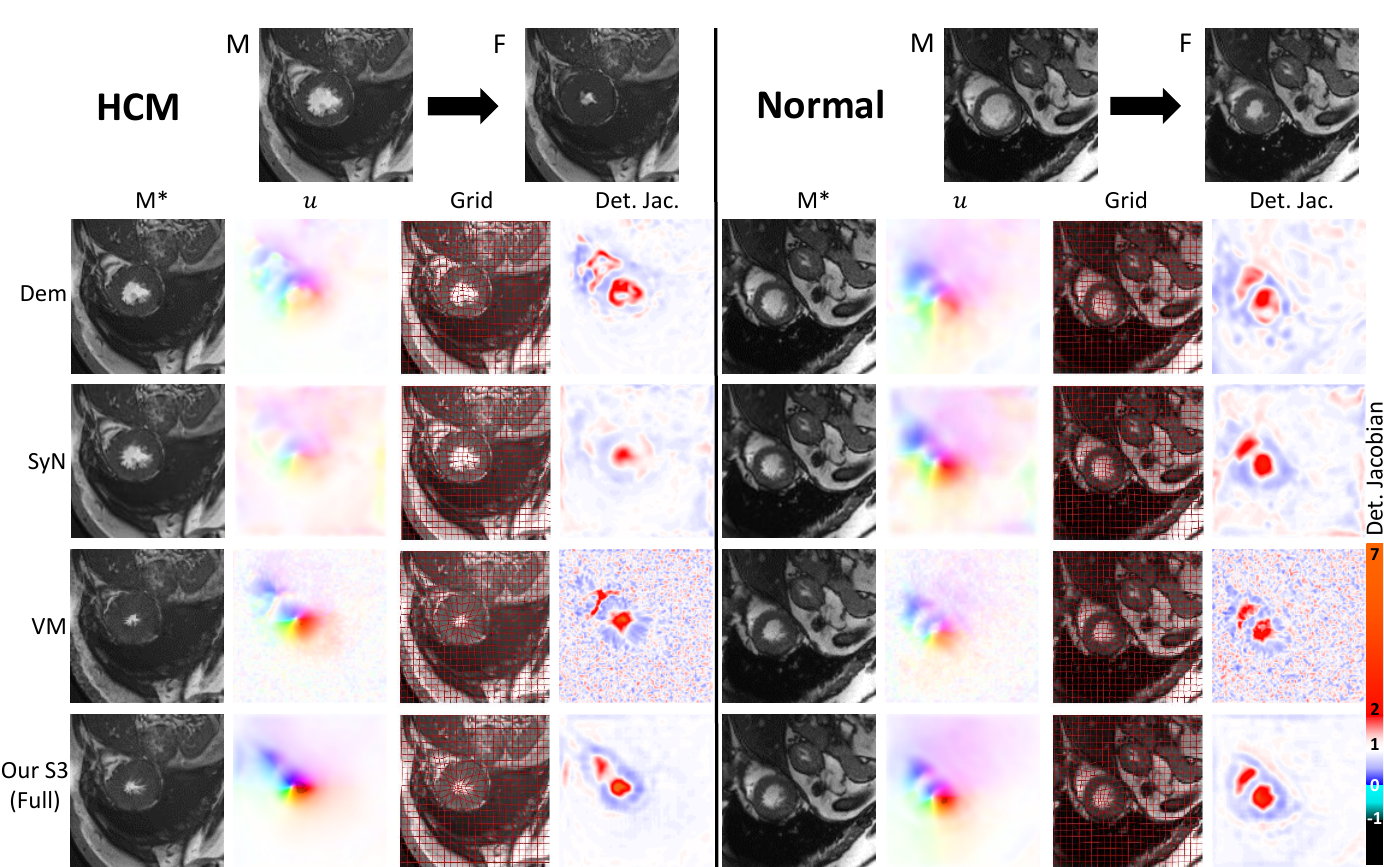}\label{reg_results}} \\
\subfloat[]{\includegraphics[trim=3 372 277 0,clip,width=1.\linewidth]{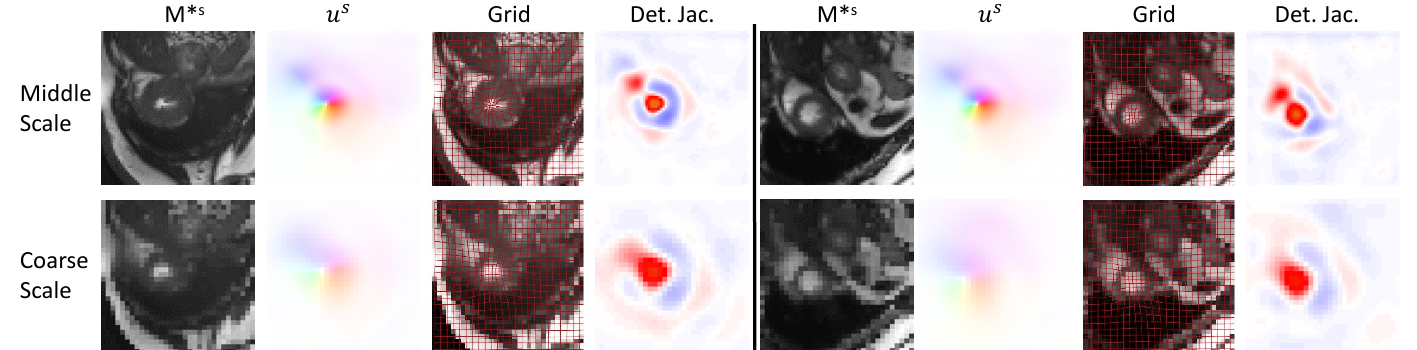}\label{multi_scaleres}}
\caption{\small{(a) Qualitative registration results showing a pathological (hypertrophy) and a normal case. Warped moving image $M^*$, \update{displacements $u$, warped moving image }with grid overlay and Jacobian determinant are shown for LCC-demons (Dem), \update{SyN}, voxelmorph (VM) and our approach using 3 scales (Our S3). (b) Middle and coarse scale predictions of our multi-scale method (Our S3).}}
\end{figure*}

We measure registration performance with the following surrogates: intensity root mean square error (RMSE), DICE score, 95\%-tile Hausdorff distance (HD in mm)\update{. To quantify deformation regularity, we show the determinant of the Jacobian qualitatively, while we also computed the mean magnitude of the gradients of the determinant of the Jacobian (Grad Det-Jac). We decided to report this second-order description of deformations to better quantify differences in smoothness among the different methods, which are not obvious by taking the mean of the determinant of the Jacobian as bigger and smaller values tend to cancel each other out.} DICE and HD scores were evaluated on the following anatomical structures: myocardium (LV-Myo) and epicardium (LV) of the left ventricle, left bloodpool (LV-BP), right ventricle (RV) and LV+RV (Fig.\ \ref{masks}).

\begin{figure}[tb]
\centering 
\includegraphics[trim=1 130 472 2,clip,width=1.\linewidth]{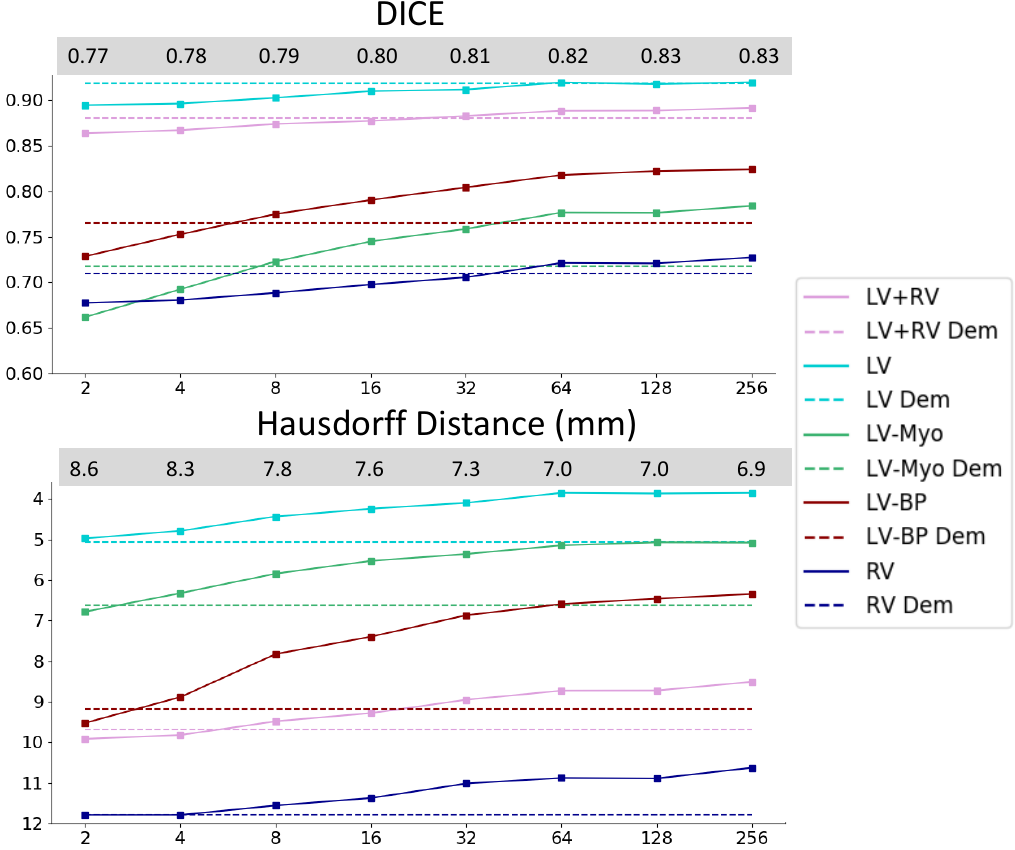}
\caption{\small{Showing the influence of the latent vector size $d$ on the registration accuracy in terms of DICE and Hausdorff distances \update{in mm} of the different anatomical structures \update{with the mean of all structures shown in the grey boxes}. The performance of the LCC-demons (Dem) is shown as reference with dashed lines.}}\label{latent_code_analysis}
\end{figure}
\begin{figure}[tb]
\centering
\includegraphics[trim=25 215 540 40,clip,width=.85\columnwidth]{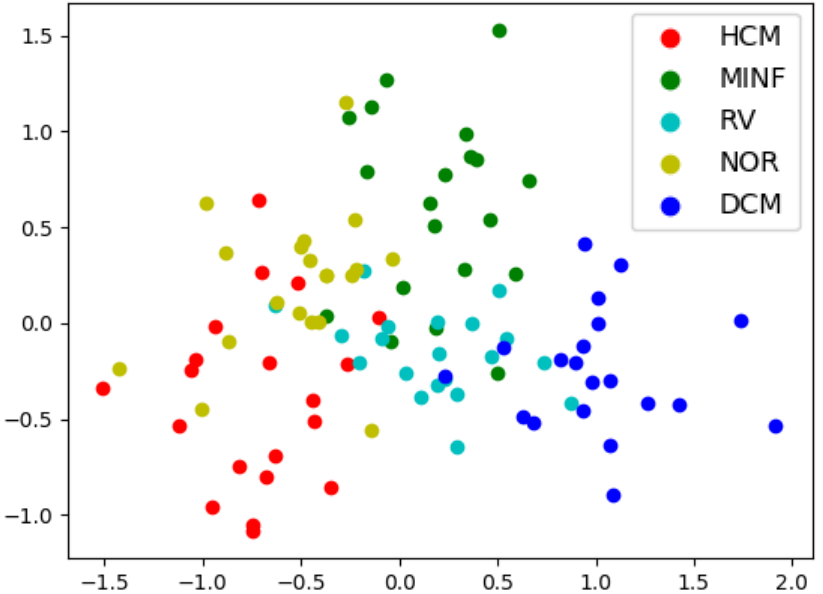}
\caption{\small{Cardiac disease distribution after projecting the latent variables $z$ of the test images on a 2-D CCA (canonical correlation analysis) space. Using an 8-D CCA and applying SVM with 10-fold cross-validation leads to a classification accuracy of 83\%}}\label{cluster}
\end{figure}
\begin{figure*}[tb]
\centering 
\includegraphics[trim=2 267 410 0,clip,width=0.70\linewidth]{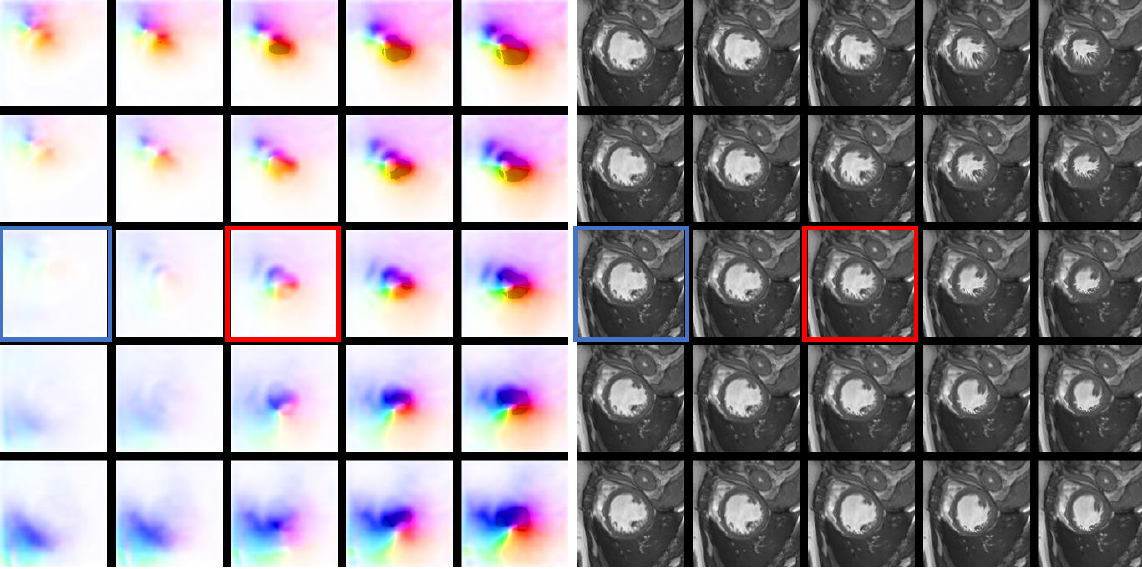}
\caption{\textcolor{black}{\small{\update{Reconstruction of simulated displacements and an accordingly warped random test image after generating $z$-codes by equally sampling along the two largest principal components within a range of $\pm2.5$ sigma around their mean values (red box). The PCA was fitted using all training $z$-codes}. The blue box indicates the image closest to the identity deformation. One can see that the horizontal eigenvalue influences large deformations while the vertical eigenvalue focuses on smaller ones, for example the right ventricle.}}}\label{manifold}
\end{figure*}

Table \ref{resultTable} shows the mean results and standard deviations of all algorithms. In terms of DICE \update{scores}, our algorithm using three scales (Our S3) shows the best performances on this dataset while the single-scale version (Our S1) performed similarly compared to the LCC-demons \update{and the SyN algorithm. Hausdorff distances were significantly improved using both of our algorithms.} Detailed registration results are shown in Fig.~\ref{cardiac_metric}. \update{Interestingly, we found that the SyN algorithm showed marginally better DICE scores than the LCC-demons which has been also reported on brain data \cite{lorenzi2013lcc}.}

Qualitative registration results of a pathological (HCM) and a healthy case (Normal) are presented in Fig.\ \ref{reg_results}\update{\footnote{Qualitative registration results for all five diseases are also presented in Fig. 10-12 (available in the supplementary files /multimedia tab).}. The warped moving image (with and wihout grid overlay) and  the determinant of the Jacobian (Det.~Jac.) are shown. Displacements} are visualized using the color encoding as typical for the optical flow in computer vision tasks. Middle and coarse scale outputs of our multi-scale method are shown in Fig.~\ref{multi_scaleres}. \update{We computed the determinant of the Jacobian using \emph{SimpleITK}\footnote{\update{\url{http://www.simpleitk.org/}}} and found that for all methods no negative values were observed on our test dataset. Compared to the other algorithms, our} approach produced smoother and more regular deformations as \update{qualitatively shown by the determinant of the Jacobian in Fig.~\ref{reg_results} and quantitatively by the significantly smaller mean gradients of the determinant of the Jacobian (Table \ref{resultTable})\update{\footnote{Visualization of the gradients of the Jacobian determinant are presented in Fig. 13 (available in the supplementary files /multimedia tab).}}. Despite the fact of being diffeomorphic}, the voxelmorph algorithm produced more irregular deformation fields compared to all other algorithms. Our single-scale approach resulted in slightly smoother deformations which is probably due to the fact that it performed less accurately in compensating large deformations. 

\begin{table}[ht]
\centering
\caption{\small{Registration performance with mean and standard deviation scores (in brackets) of RMSE, DICE, Hausdorff Distance (HD \update{in mm}) and the mean gradient of the determinant of Jacobians (Grad Det-Jac, $\times 10^{-2}$) comparing the undeformed case (Undef), LCC-demons (Dem), \update{SyN}, voxelmorph (VM) and our method.}}\label{resultTable}
\begin{tabular}{l|rrrrr}
Method & RMSE & DICE & HD & Grad Det-Jac \\
\hline
Undef & 0.37 (0.17) & 0.707 (0.145) & 10.1 (2.2) & --\\ 
Dem & 0.29 (0.16) & 0.799 (0.096) & 8.3 (2.7) & 2.9 (1.0) \\ 
\update{SyN} & \update{0.32 (0.16)} &\update{0.801 (0.091)} & \update{8.1 (3.6)} &\update{3.4 (0.5)} \\
VM & \bf{0.24 (0.08)} & 0.790 (0.096) & 8.4 (2.6) & 9.2 (0.5) \\ 
Our S1 & 0.31 (0.15) & 0.797 (0.093) & 7.9 (2.6) & \bf{1.2 (0.3)} \\ 
Our S3 & 0.30 (0.14) & \bf{0.812 (0.085)} & \bf{7.3 (2.7)} & 1.4 (0.3)\\ 
\noalign{\smallskip}
\hline
\end{tabular}
\end{table}

\update{We applied the Wilcoxon signed-rank test with $p<0.001$ to evaluate statistical significance of the differences in the results of Fig.~\ref{cardiac_metric}. This method is chosen as a paired test without the assumption of normal distributions. For all metrics, the results of our multi-scale algorithm (Our S3) showed significant differences compared to the results of all other methods (including Our S1). With respect to our single-scale algorithm (Our S1), only the differences in DICE scores were not statistically significant in comparison with the LCC-demons (Dem).} 

\update{Note, that higher DICE and HD scores can be achieved by choosing a higher latent dimensionality (cf.\ Experiment \ref{lat_size}), which however comes at the cost of a more complex encoding space, making analysis tasks more difficult.} We also tested the first version of voxelmorph \cite{balakrishnan2018unsupervised} on our dataset. We chose to show the results of the latest version \cite{dalca2018unsupervised} due to the fact that this version is diffeomorphic and that its DICE and HD results were better (cf.\ \cite{krebs2018unsupervised}).

\subsubsection{Deformation encoding}
For evaluating the learned latent space, we investigated (a) the effects of the size of the latent vector on the registration accuracy, (b) the structure of the encoded space by visualizing the distribution of cardiac diseases and showing \update{simulated deformations along the two main axes of variations} and (c) we applied our framework on deformation transport and compare its performance with a state-of-the-art algorithm. 

\paragraph{Latent Vector Size}\label{lat_size} In Fig.~\ref{latent_code_analysis} we analyzed the influence of the size of the latent code vector with respect to registration accuracy in terms of DICE and HD scores. With a relatively small latent size of $d=8$, competitive accuracy is achieved. With an increasing dimensionality, performance increases but reaches a plateau eventually. This behavior is expected, since CVAEs tend to ignore components if the dimensionality of the latent space is too high \cite{kingma2014semi}. For the cardiac use case, we chose $d=32$ components as a trade-off between accuracy and latent variable size.

\paragraph{Disease Distribution and \update{Generative Latent Space}} In this experiment, we used disease information and encoded z-codes of the test images to visualize \update{the} learned latent space. Using linear CCA (canonical correlation analysis), we projected the $z$-codes (32-D) to a 2-D space by using the two most discriminative CCA components. Fig.~\ref{cluster} shows that the 100 test sets are clustered by classes in this space. Taking the 8 most discriminative CCA components into account, the classification accuracy of the five classes is \textbf{83\%} with 10-fold cross-validation using support vector machines (SVM). 
In a second experiment, we applied principal component analysis (PCA) on the $z$-codes of the \update{training dataset. We simulated deformations by sampling equally distributed values in the range of $\pm2.5$ standard deviations of the two largest principal components} and extracting the $z$-codes through inverse projections. Fig.~\ref{manifold} shows \update{reconstructed displacements} and deformed images when applying these \update{generated} $z$-codes on a random test image. One can see the different influences of the two eigenvalues. The first eigenvalue (horizontal) focuses on large deformations while the second one focuses on smaller deformations as the right ventricle. The results of these two experiments which are solely based on applying simple linear transformations, suggest that deformations that are mapped close to each other in the deformation latent space have similar characteristics.

\paragraph{Deformation Transport}
Pathological deformations can be transported to healthy subjects by deforming the healthy ES frames using pathological ED-ES deformations\update{. Our framework allows for deformation transport by first registering the ED-ES frames of a given pathological case (Step 1), which we call \emph{prediction} in this experiment. Secondly, we use the $z$-codes from these predictions along with the ED frame of a healthy subject to transport the deformations (Step 2, cf.\ Fig.~\ref{decoder}). Note, that this procedure does not require} any inter-subject registrations. 

We compare our approach with the pole-ladder algorithm (PL \cite{lorenzi2014efficient}). All intra- and inter-subject registrations required by the pole ladder were performed using the LCC-demons \cite{lorenzi2013lcc}. For the inter-subject pairs, we aligned the test data with respect to the center of mass of the provided segmentation and rotated the images manually for rigid alignment. This alignment step was done only for the pole ladder experiment\update{\footnote{The pipeline for the parallel transport experiment using the pole ladder algorithm is presented in Fig. 14 (available in the supplementary files /multimedia tab).}}.

\begin{figure*}[tb]
\centering 
\includegraphics[trim=1 226 230 2,clip,width=1.\linewidth]{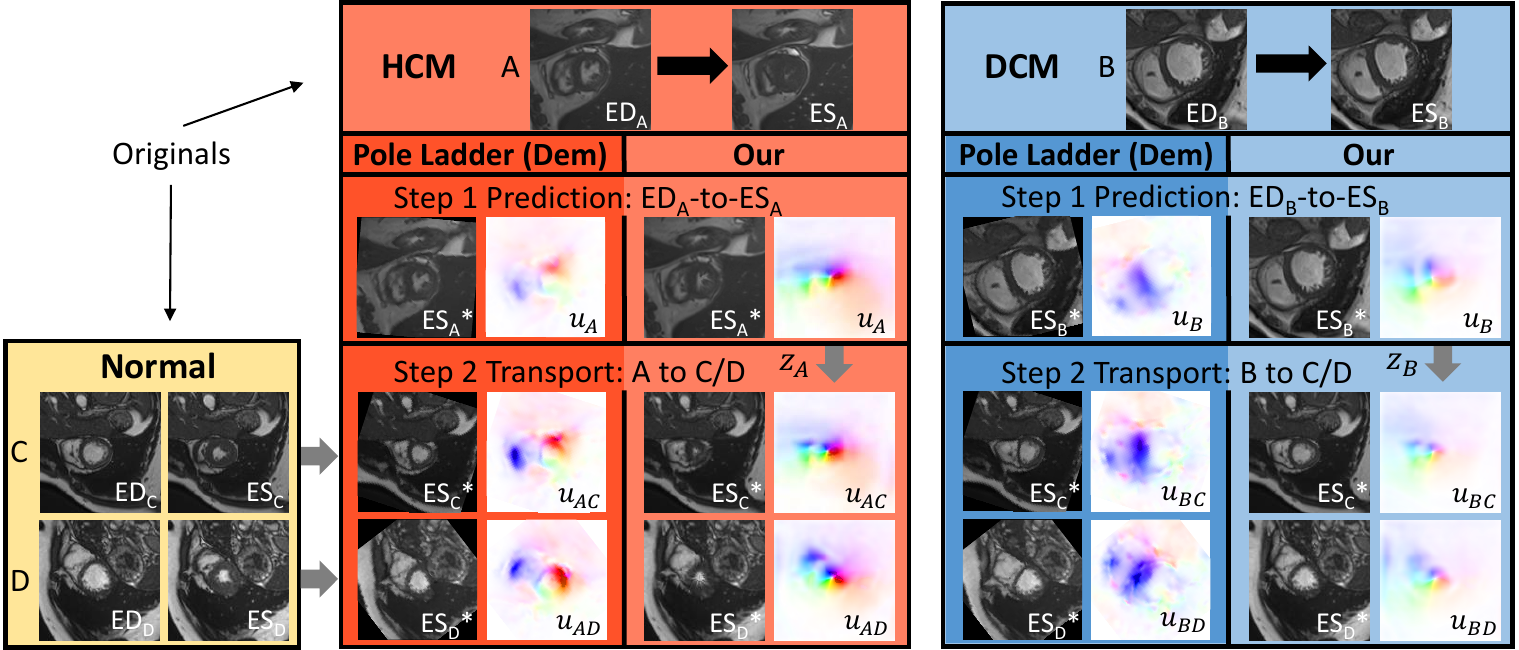}
\caption{\small{Transport pathological deformation predictions (\update{Step 1,} hypertrophy HCM, myopathy DCM) to healthy (Normal) subjects by using the pole ladder (with LCC-demons) and our probabilistic method \update{(Step 2)}. Note that the pole ladder algorithm requires the registration between pathological and normal subjects while our approach is able to rotate and translate deformations encoded in the latent space $z$ automatically.}}\label{transport}
\end{figure*}
Qualitative results are shown in Fig.~\ref{transport} where the predicted deformations of one hypertrophy (A, HCM) and one cardiomyopathy (B, DCM) case \update{(step 1)} were transported to two healthy \update{(Normal) subjects (step 2,} targets C and D). Note that our algorithm automatically determines orientation and location of the heart. In Table \ref{transportEFTable}, \update{we evaluated the average ejection fraction (EF) of the ED-ES deformation prediction of the pathologies (step 1) and the average EF after transport to normal subjects (step 2). Hereby, we assume that EFs, as a relative measure, stay similar after successful transport (such that the absolute difference, EF step 1 - EF step 2, is small).} 
\begin{table}[h]
\centering
\caption{\small{Mean Ejection fraction (EF in \% with standard deviation in parentheses) of \update{pathological deformation predictions (Step 1) should stay similar to the mean EF after the transport to healthy/normal subjects (Step 2). Our algorithm shows smaller absolute differences compared to the pole ladder (PL).}}}\label{transportEFTable}
\begin{tabular}{l|rr|rr|rr}
 & \multicolumn{2}{c}{Step 1: Prediction} & \multicolumn{2}{c}{Step 2: Transport} & \multicolumn{2}{c}{\update{Difference}} \\
Path. & PL (Dem) & Our & PL (Dem) & Our & \update{PL} & \update{Our} \\
\hline
HCM & 29.4 (6) & 44.1 (7) & 35.5 (8) & 38.6 (13) & \update{6.1} & \update{\bf{5.5}}\\ 
DCM & 10.8 (2) & 12.7 (7) & 16.7 (4) & 13.9 (7) & \update{5.9} & \update{\bf{1.2}}\\ 
\noalign{\smallskip}
\hline
\end{tabular}
\end{table}
The table shows the average of transporting 5 HCM and 5 DCM cases to 20 normal cases (200 transports).
\update{For our algorithm, the absolute differences in EFs are much smaller for DCM cases and similarly close in HCM cases in comparison to the pole ladder. }\update{All test subjects were not used during training. The EF is computed based on the segmentation masks (warped with the resulting deformation fields). Besides, it can be seen, that predictions done by the demons are underestimating the EFs for HCM cases which should be $>$40\% according to the ACDC data set specifications.} 

\section{Discussion and Conclusions}
We presented an unsupervised multi-scale deformable registration approach that learns a low-dimensional probabilistic deformation model. Our method not only allows \update{the accurate registration of} two images but also \update{the analysis of} deformations. The framework is generative, as it is able to simulate deformations \update{given only one image. Furthermore, it provides a novel way of deformation transport} from one subject to another by using its probabilistic encoding. In the latent space, similar deformations are close to each other. \update{The} method enables \update{the addition of} a regularization term which leads to arbitrarily smooth deformations that are diffeomorphic by using an exponentiation layer for stationary velocity fields. The multi-scale approach, providing velocities, deformation fields and warped images in different scales, leads to improved registration results and a more controlled training procedure compared to a single-scale approach. 

We evaluated the approach on end-diastole to end-systole cardiac cine-MRI registration \update{and compared registration performance in terms of RMSE, DICE and Hausdorff distances to two popular algorithms  \cite{lorenzi2013lcc,avants2008symmetric} and a learning-based method \cite{dalca2018unsupervised}}, which are all diffeomorphic. \update{While the performance of our single-scale approach was comparable to the LCC-demons and the SyN algorithm, our multi-scale approach (using 32 latent dimensions) showed statistically significant improvements in terms of registration accuracy. Generally, our approach produced more regular deformation fields, which are significantly smoother than the DL-based algorithm.} Using our method with a non-generative U-net style network \cite{ronneberger2015u} without a deformation encoding performed similarly compared to the proposed generative model. Adding supervised information such as segmentation masks in the training procedure as in \cite{hu2018weakly,fan2018birnet} led to a marginal increase in terms of registration performance ($\sim$1-2\% in DICE scores), \update{so we decided that the performance} gain is not large enough in order to justify the higher training complexity. Theoretically, our method \update{allows measurement of} registration uncertainty as proposed in \cite{dalca2018unsupervised} which we did not further investigate in this work. 

The analysis of the deformation encoding showed that the latent space projects similar deformations close to each other such that diseases can be clustered. Disease classification could be potentially enforced in a supervised way as in \cite{biffi2018learning}. Furthermore, our method showed comparable quantitative and qualitative results in transporting deformations with respect to a state-of-the-art algorithm which requires the difficult step of inter-subject registration that our algorithm does not need. 

It is arguable if the simple assumption of a multivariate Gaussian is the right choice for the prior of the latent space (Eq.~\ref{p_z}). Possible other assumptions such as a mixture of Gaussians are subject to future work. The authors think that the promising results of the learned probabilistic deformation model could be also applicable for other tasks such as evaluating disease progression in longitudinal studies or detecting abnormalities in subject-to-template registration. An open question is how the optimal size of the latent vector changes in different applications. In future work, we plan to further explore generative models for learning probabilistic deformation models.

\section*{Acknowledgments}
Data used in preparation of this article were obtained from the EU FP7-funded project MD-Paedigree and the ACDC STACOM challenge 2017 \cite{bernard2018deep}. 
This work was supported by the grant AAP Sant\'e 06 2017-260 DGA-DSH, and by the Inria Sophia Antipolis - M{\'e}diterran{\'e}e, "NEF" computation cluster. The authors would like to thank \update{Xavier Pennec for the insightful discussions and} Adrian Dalca for the help with the Voxelmorph \cite{dalca2018unsupervised} experiments. 

\noindent \textbf{Disclaimer: } This feature is based on research, and is not commercially available. Due to regulatory reasons its future availability cannot be guaranteed.

\ifCLASSOPTIONcaptionsoff
  \newpage
\fi

\bibliography{IEEEabrv.bib,strucDeformations.bib}

\setcounter{figure}{9}
\begin{figure*}[!ht]
\centering 
\subfloat[]{\includegraphics[width=1.0\linewidth]{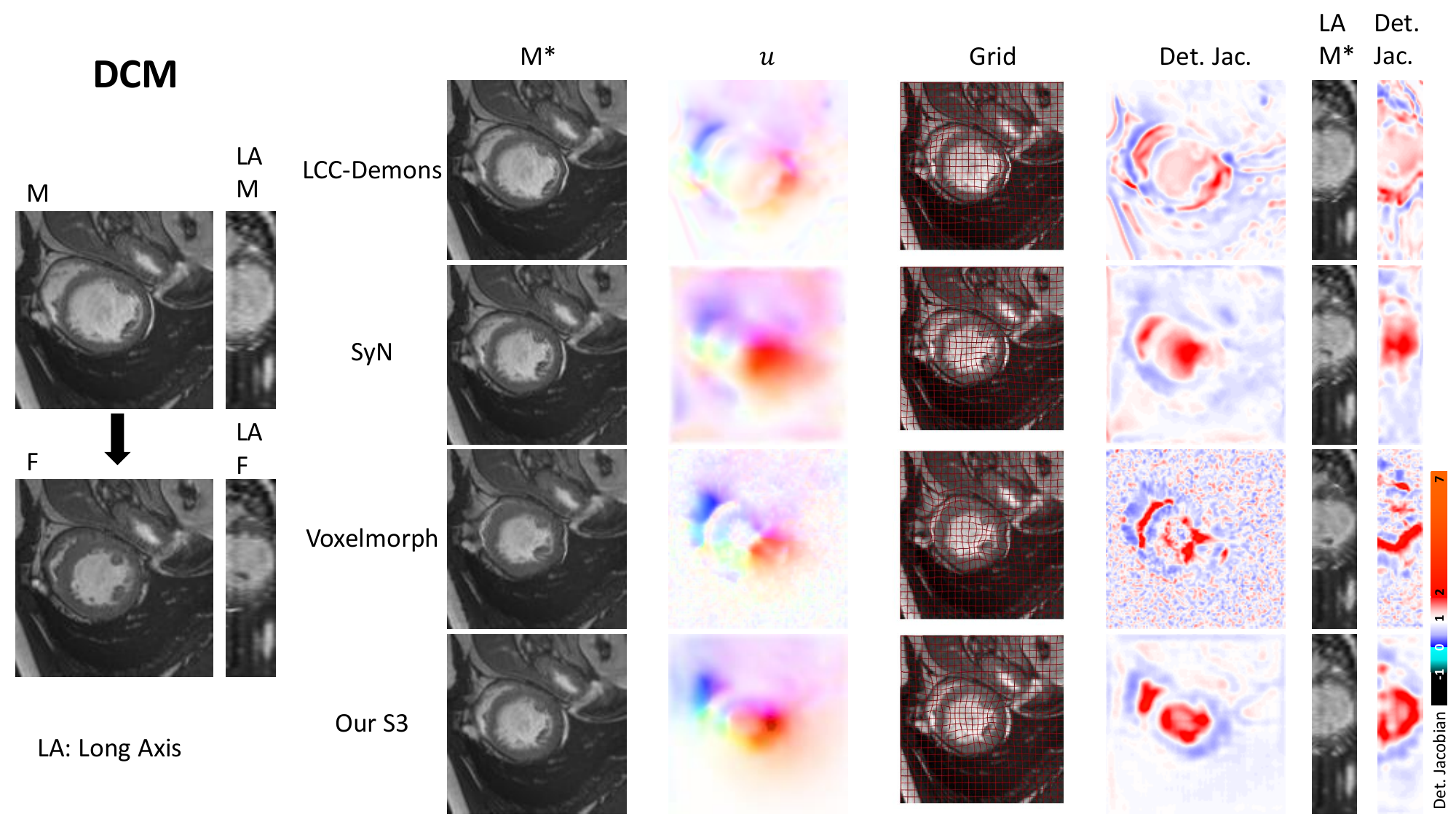}\label{sub_0}}\\
\subfloat[]{\includegraphics[width=1.0\linewidth]{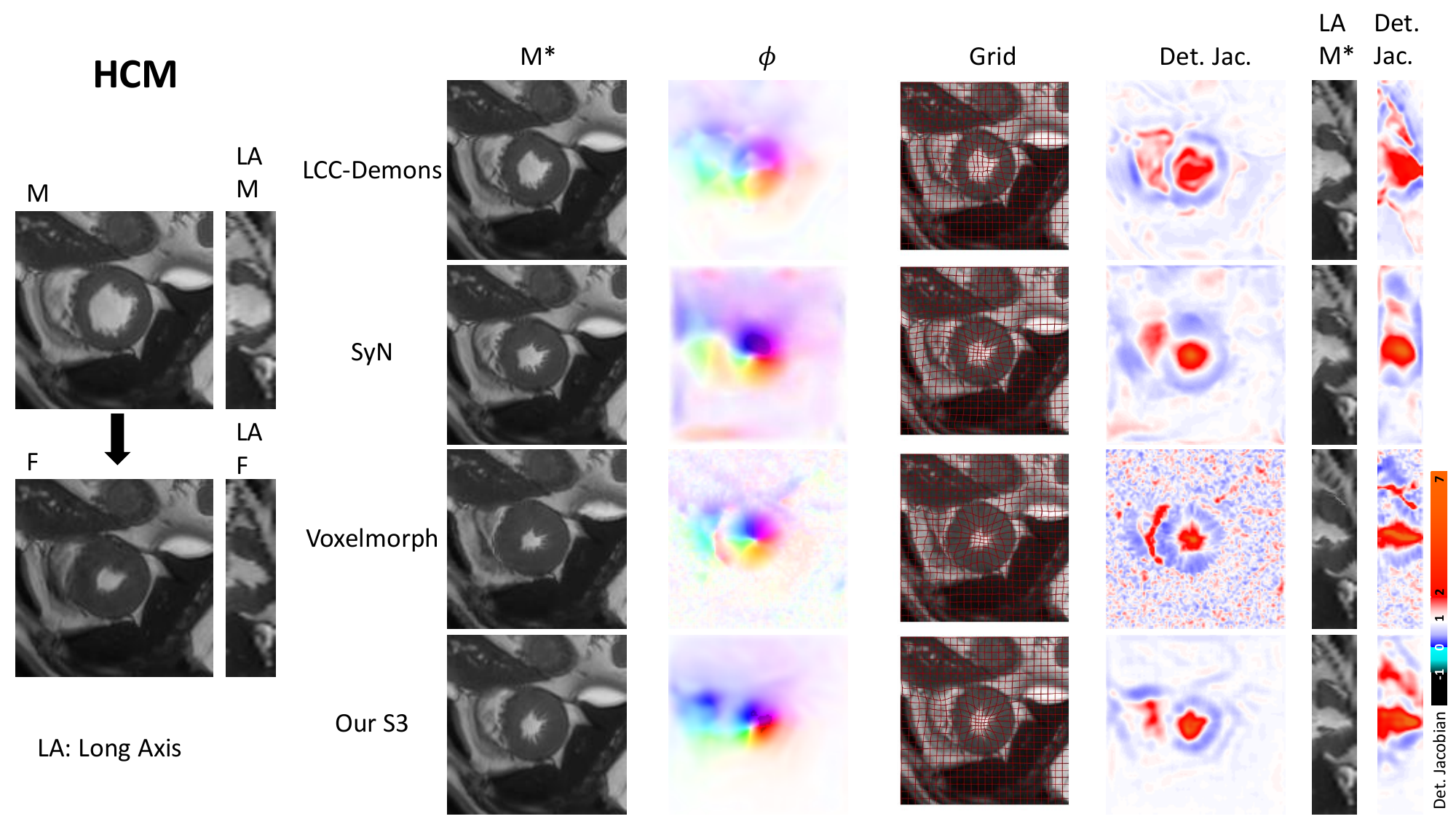}\label{sub_1}}
\caption{\small{Qualitative registration results showing a dilated cardiomyopathy  (DCM) and a hypertrophic cardiomyopathy (HCM) case. Warped moving image $M^*$, displacements $u$, warped moving image with grid overlay and Jacobian determinant are shown for LCC-demons (Dem), SyN, voxelmorph (VM) and our approach using 3 scales (Our S3).}}
\end{figure*}

\begin{figure*}[!ht]
\centering 
\subfloat[]{\includegraphics[width=1.0\linewidth]{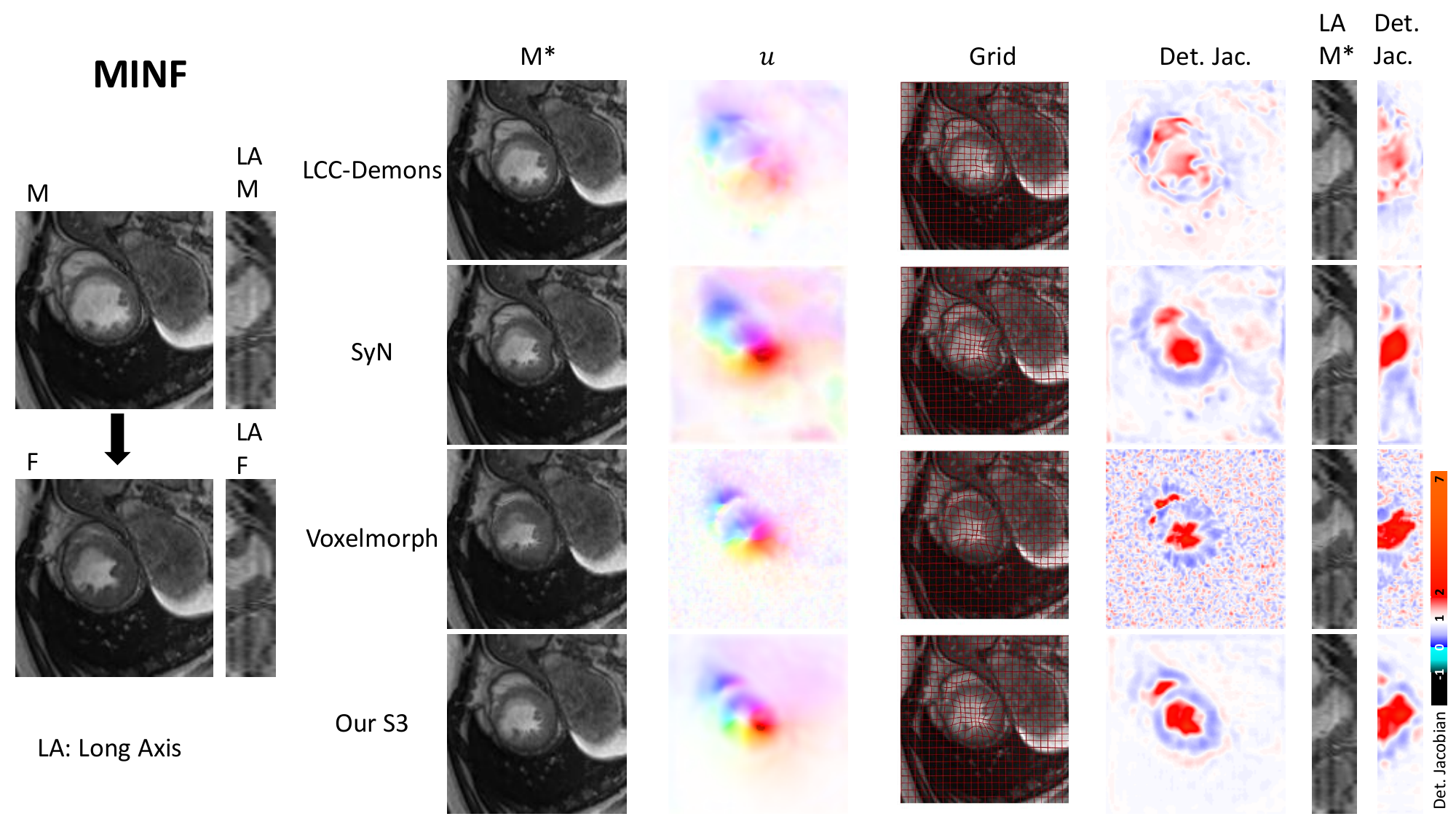}\label{sub_2}}\\
\subfloat[]{\includegraphics[width=1.0\linewidth]{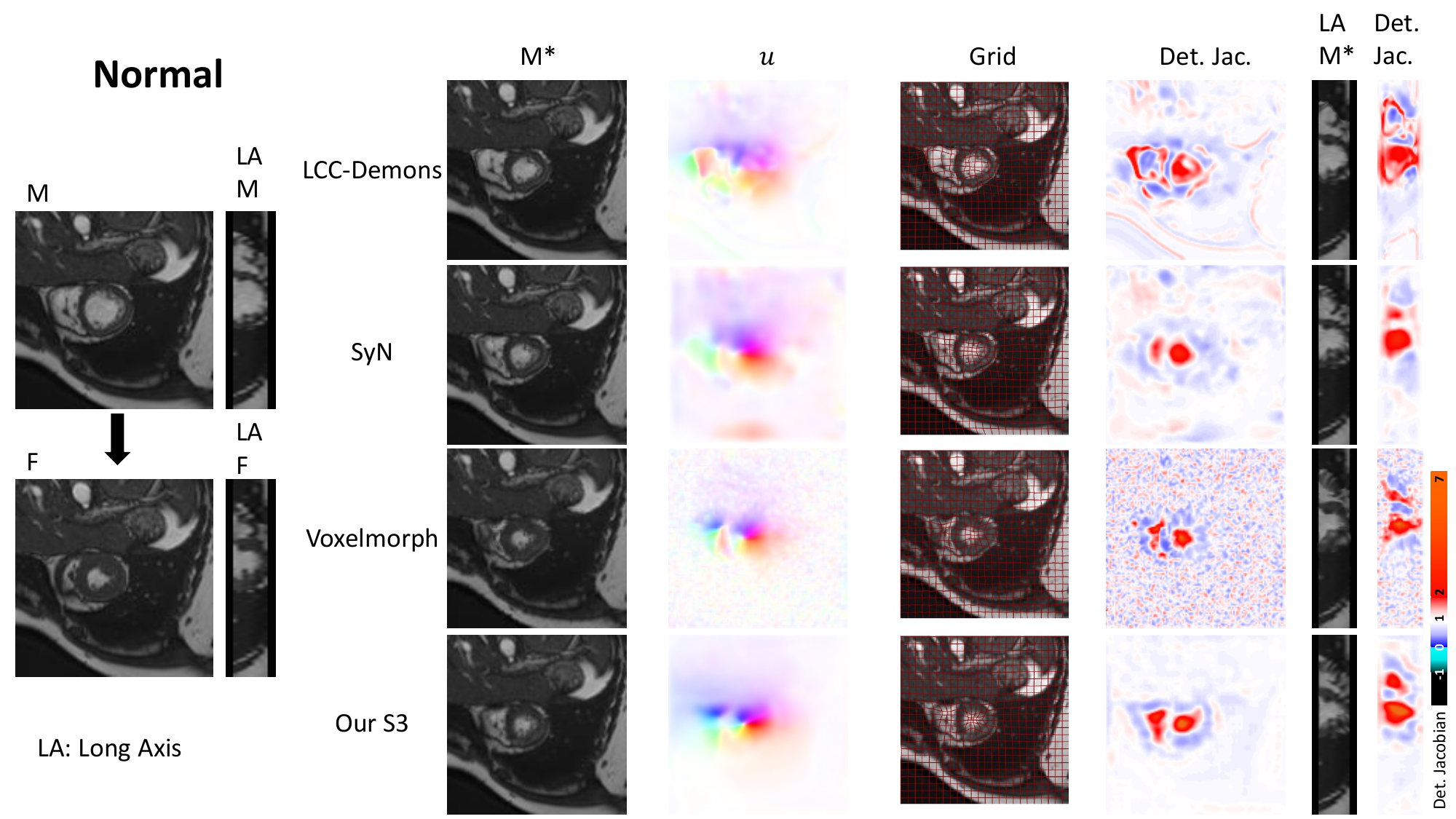}\label{sub_3}}
\caption{\small{Qualitative registration results showing a myocardical infarction (MINF) and healthy (Normal) case. Warped moving image $M^*$, displacements $u$, warped moving image with grid overlay and Jacobian determinant are shown for LCC-demons (Dem), SyN, voxelmorph (VM) and our approach using 3 scales (Our S3).}}
\end{figure*}

\begin{figure*}[!ht]
\centering 
\includegraphics[width=1.0\linewidth]{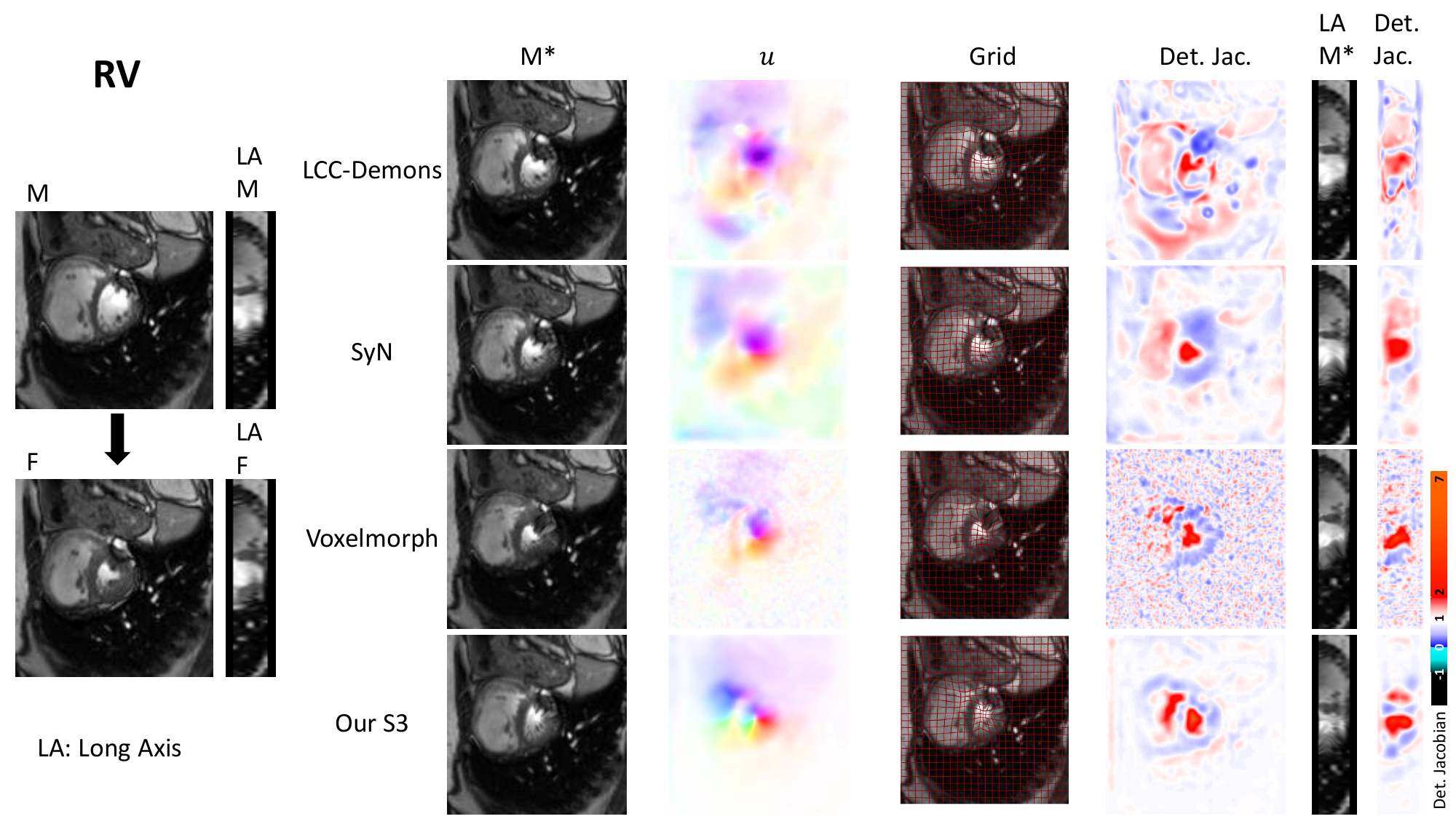}\label{sub_4}
\caption{\small{Qualitative registration results showing an abnormal right ventricle case (RV). Warped moving image $M^*$, displacements $u$, warped moving image with grid overlay and Jacobian determinant are shown for LCC-demons (Dem), SyN, voxelmorph (VM) and our approach using 3 scales (Our S3).}}
\end{figure*}

\begin{figure*}[!ht]
\centering 
\includegraphics[trim=0 260 0 65,clip,width=1.0\linewidth]{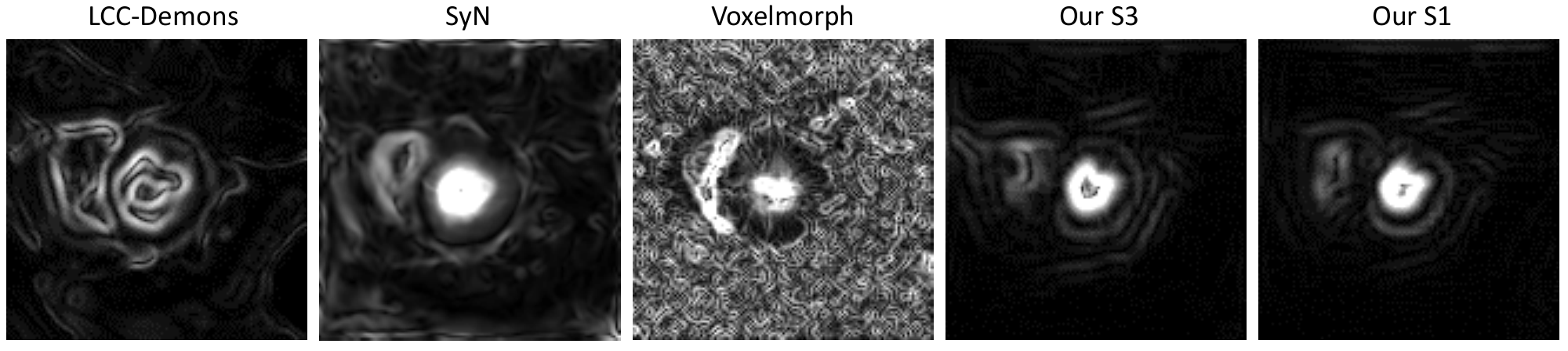}\label{sub_5}
\caption{\small{The gradient of the determinant of the Jacobian of a random test case for LCC-demons (Dem), SyN, voxelmorph (VM) and our approach using 1 and respectively 3 scales (Our S1, Our S3). Our single-scale approach shows the most regular deformation.}}
\end{figure*}

\begin{figure*}[!ht]
\centering 
\subfloat[]{\includegraphics[trim=0 20 0 65,clip,width=1.0\linewidth]{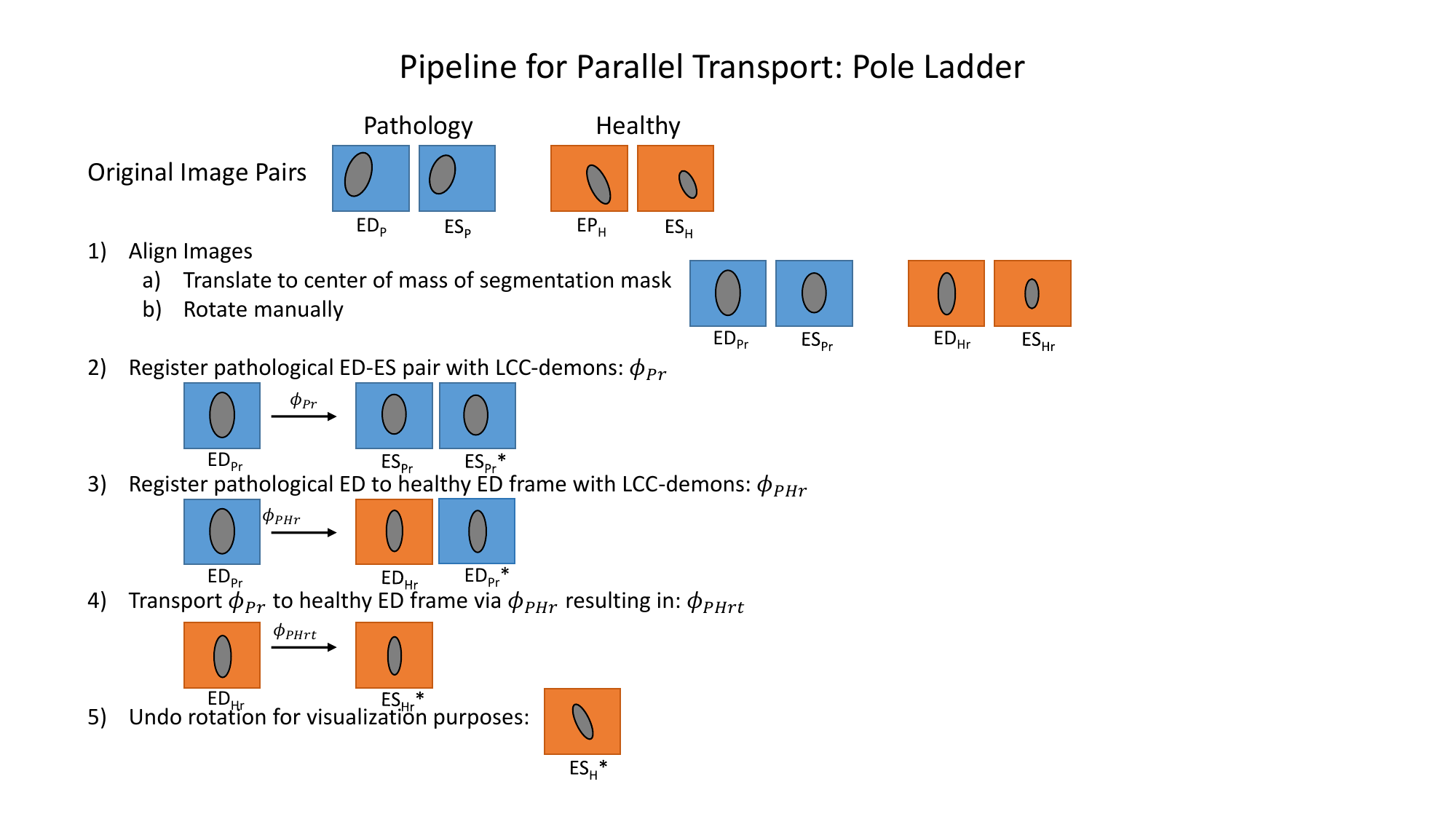}\label{sub_6}}\\
\subfloat[]{\includegraphics[trim=0 20 0 65,clip,width=1.0\linewidth]{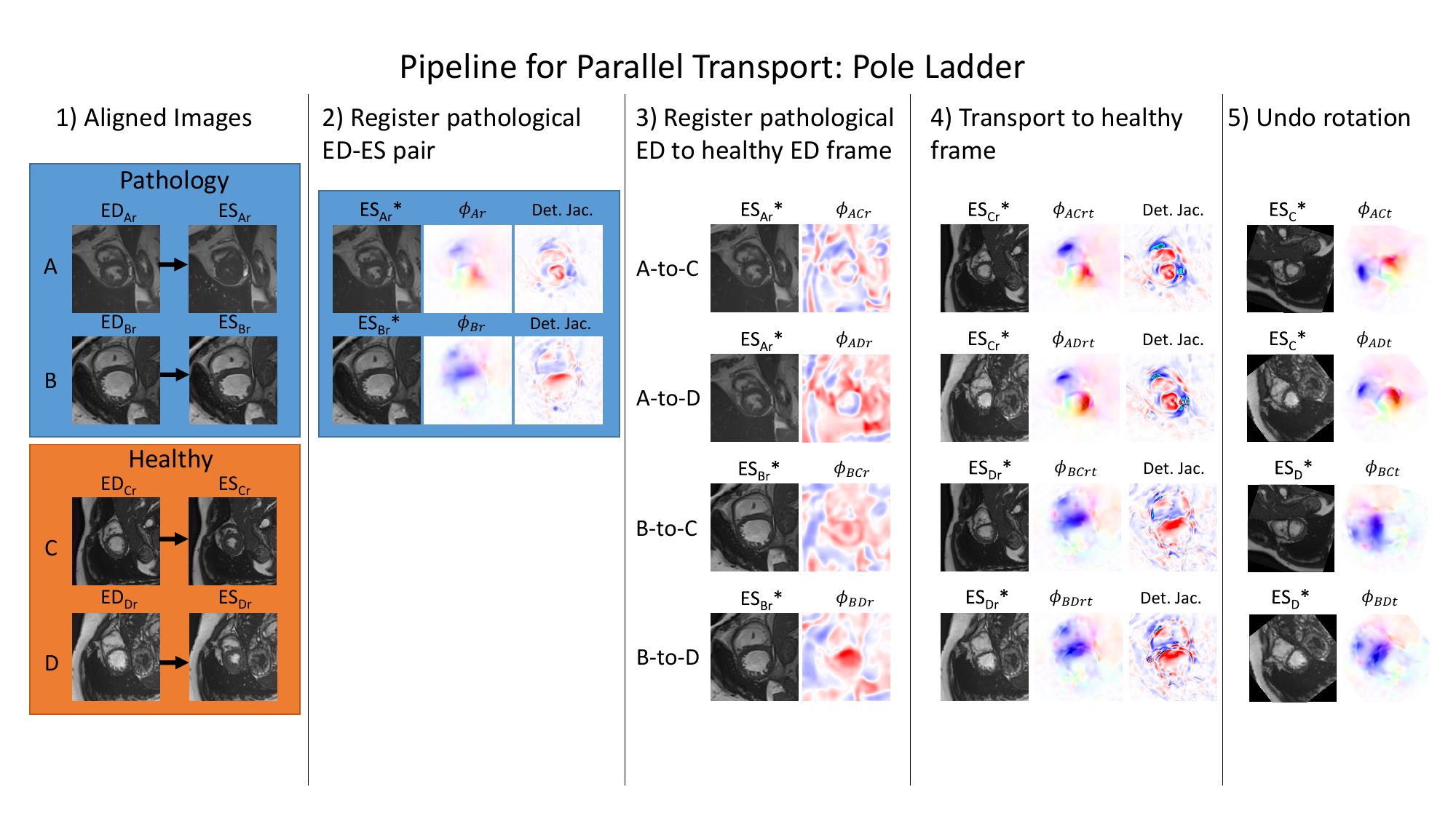}\label{sub_7}}
\caption{\small{(a) Symbolic pipeline for the parallel transport experiment using the pole ladder approach. (b) Visualization of all pipeline steps for one example.}}
\end{figure*}

\end{document}